\pdfoutput=1
\documentclass[11pt]{article}

\usepackage{acl}

\usepackage{times}
\usepackage{latexsym}
\usepackage{amsfonts}

\usepackage[T1]{fontenc}
\usepackage[utf8]{inputenc}

\usepackage{microtype}

\usepackage{inconsolata}

\usepackage{caption}
\usepackage{subcaption}
\usepackage{url}
\usepackage{amsmath}
\usepackage{multirow}
\usepackage{float}
\usepackage{amssymb}
\usepackage{comment}
\usepackage{linguex}
\usepackage{booktabs}

\usepackage{graphicx}
\usepackage{colortbl}
\usepackage{soul}

\definecolor{darkcerulean}{rgb}{0.03, 0.27, 0.49}

\definecolor{pine}{rgb}{0.10, 0.40, 0.30}

\definecolor{tan}{rgb}{0.69, 0.4, 0.0}

\definecolor{LightGrayText}{HTML}{8A8A8A}

\newcommand{\lightgraytext}[1]{\textcolor{LightGrayText}{#1}}

\usepackage[capitalize,noabbrev]{cleveref}
\crefname{section}{\S}{\S\S}
\crefname{table}{Tab.}{Tab.}
\crefname{figure}{Fig.}{Figs.}
\crefname{algorithm}{Alg.}{}
\crefname{equation}{Eq.}{Eq.}
\crefname{appendix}{App.}{App.}
\crefname{theorem}{Theorem}{}
\crefname{restatableTheorem}{Theorem}{}
\crefname{prop}{Proposition}{}
\crefname{definition}{Def.}{}
\crefname{cor}{Corollary}{}
\crefname{observation}{Observation}{}
\crefname{assumption}{Assumption}{}
\crefname{hyp}{Hyp.}{Hypotheses}
\crefformat{section}{\S#2#1#3}
\crefname{namedtheorem}{Hyp.}{Hypotheses}

\title{Patterns of Priming in Production: 
Lexical, Semantic and Structural Alignment in Language Model Generation
}

\author{Giulia Pucci$^{\heartsuit}$ \quad 
Ruizhe Li$^{\heartsuit\mathbb{B}}$ \quad 
Arabella Sinclair$^{\heartsuit\Cup}$\\
  School of Natural and Computing Science, University of Aberdeen$^{\heartsuit}$ \\
  School of Computer Science, University of Birmingham$^{\mathbb{B}}$ \\
  Department of Information Studies, University College London$^{\Cup}$ \\
  \texttt{giulia.pucci@abdn.ac.uk, r.li.7@bham.ac.uk, arabella.sinclair@ucl.ac.uk}
  \\}

\begin{document}
\maketitle
\begin{abstract}
This paper investigates structural priming in language model (LM) production, examining how preceding structural context influences sentence completion.
While prior work has demonstrated priming effects in \textit{comprehension} of structural alternations, it remained unclear whether these persist in \textit{production}, where, when generating, an LM samples from many possible continuations at each step. 
We address this question through a series of controlled sentence-completion experiments on dative constructions.
In line with prior work, we find that LMs are susceptible to structural priming, particularly in sentences that are semantically coherent.
In terms of priming magnitude, we find that while there is a greater \textit{relative} increase of double-object datives against our baselines, in line with inverse frequency effects, 
there is a larger absolute increase in prepositional-objects, the more frequently produced construction.
Finally, we not only observe that structural priming is boosted by lexico-semantic coherence, but that structurally primed completions display greater levels of lexico-semantic repetition. 
Taken together, our evidence supports the view that structural priming in LMs operates across multiple levels of linguistic representation, facilitating, and facilitated by syntactic, lexical, and semantic alignment.\footnote{
Code and data can be found at
\url{https://github.com/the-context-lab/primedproduction}
}
\end{abstract}

% --------------------------------------------
\section{Introduction}
\label{sec:intro}
% --------------------------------------------
Structural priming in production refers to the phenomenon whereby a speaker, having recently been exposed to a specific syntactic structure, is more likely to re-use that structure in subsequent utterances~\citep{bock1986syntactic}. 
This phenomenon is well-documented in human production \citep{mahowald2016meta}, but remains less clearly understood in Language Models (LMs). While recent studies have demonstrated structural priming in LLM \textit{comprehension}, that is, lower surprisal for previously encountered syntactic structures \citep{sinclair2022structural, jumelet-etal-2024-language}, less is known about how these effects manifest in \textit{production}, which we make our focus. 
In humans, effects of priming show some degree of parity across both comprehension and production~\citep{TooleyKM}, with priming effects amplified by lexical overlap, and sensitive to discourse structure~\citep[e.g.]{Ziegler16032019,pickering1998representation}. While it has been demonstrated that LMs display a similar lexical boost effect when measuring their comprehension behaviour~\citep{jumelet-etal-2024-language}---arguably this prediction difference should also affect their behaviour when examined in generation settings. 

Evaluating model behaviour in a production setting allows us to examine whether structural priming produces broader downstream effects on sentence generation, providing a clearer understanding of how priming may emerge and accumulate across multiple linguistic levels. We predict LMs will exhibit facilitative effects of syntactically congruent contexts on lexico-semantic production choices, consistent with facilitative processing effects observed in human language production~\citep[e.g.][]{wright1984lexical}.

{
\small

\ex. The woman gave \hfill \emph{Prime Sentence}
\a. [the man]$_{NP}$ \hspace{.7 mm}[the book]$_{NP}$\hfill \emph{DO}
\b. [the book]$_{NP}$ \hspace{.7 mm}[to the man]$_{PP}$\hfill \emph{PO}

\ex. \hfill \emph{Target Sentence Stem}
\a. The lawyer saved the ...\hfill\emph{unrelated}
\b. The \textbf{man} saved the ...\hfill\emph{coherent}

}

Using a constrained sentence-completion task taking dative constructions as a case study, we seek to investigate LM primed production behaviour---understanding how prior context influences generated structure. Our experimental design was inspired by~\citet{bock1986syntactic}, but adapted to a primed sentence completion task more similar to that of~\citet{pickering1998representation,corley2002syntactic} in order to isolate the specific effects we are interested in. Given a prime (1a) we expect a given target stem (2a) to be completed in syntactically congruent structure. We expect a target stem that is lexico-semantically congruent with the prime (2b) to exhibit a greater likelihood of structural congruence due to links between priming and discourse structure~\citep[e.g.][who show priming is sensitive to discourse focus]{Ziegler16032019}, as well as findings of lexical repetition boosting priming~\citep{segaert2013syntactic}.

We construct a controlled set of experimental stimuli, extract completions from a range of LMs, and create and validate an experimental pipeline for LM completion extraction and automatic annotation to answer the following research questions: 

\begin{description}
    \item[RQ1]  Do LMs exhibit \textbf{structural priming in production}? \textit{We characterise priming as the greater incidence of matching structures in primed completions than with a) no prime or b) the alternative construction.}
    \item[RQ2] Does \textbf{semantic coherence} between a prime \textit{(1a)} and a target stem \textit{(2b)} boost structural priming? \textit{Measured as a) absolute  difference, and b) relative proportion increase.}
    \item[RQ3] Is structural priming accompanied by \textbf{lexico-semantic alignment}? \textit{ Measured by the difference in semantic similarity and lexical overlap between primed vs. unprimed completions.}
\end{description}
We observe that in a baseline no-context completion setting, LMs rarely produce dative structures, but consistently produce greater levels of PO completions. In-keeping with our main hypotheses, we firstly observe that LMs do demonstrate priming behaviour in production - with greater levels of dative constructions following structurally congruent context than in an empty, non-dative, or dative context. Secondly, we observe that semantic coherence between prime and target stem results in greater incidence of structurally primed completions. Finally, we observe that the lexico-semantic alignment of a completion to the context is greater in structurally primed settings---which, while not causal, indicates the potential of a bidirectional interaction between structural and lexico-semantic priming mechanisms during generation.

% \section{Background}
\section{Related Work}

\paragraph{Priming in Humans}
Priming effects have been found to occur across various linguistic levels, including lexical~\citep{Bock1986}, semantic~\citep{mcnamara2005semantic} and syntactic~\cite{bock1986syntactic}.
% \as{gp: add  a citation for evidence of phonetic and lexical please}
These effects are influenced by factors such as animacy ~\citep{Buckle} and ambiguity avoidance \cite{Zehenter2021}. Importantly, priming at one level can amplify effects across others~\citep[e.g. lexical repetition boosts structural priming][]{segaert2013syntactic}.
Priming in production is often measured through repetition of syntactic structures in recently seen sentences~\citep{bock1986syntactic} or adjacent utterances~\citep{Reitter2006a}. These repetitions can also take the form of morphosyntactic constructions~\citep[e.g.][]{sinclair2021construction}. 
Priming, alignment and repetition in humans are thought to facilitate comprehension \cite{pickering2004}, play a role in language acquisition~\citep{sinclair-etal-2018-ability,bala-etal-2026-pragmatic},
and can be reliably observed in human corpora~\cite{Gries2004,bresnanpredicting}.
Local repetition can also be ascribed to efficient processing mechanisms~\citep{giulianelli2022construction,yee-etal-2024-efficiency}, where residual activation facilitates retrieval of recently encountered material~\citep[e.g.][]{lexical1984}, or as a result of implicit learning~\citep{chang2000structural} whereby speakers' recent updates to their LM influence production.

\paragraph{Priming in LMs}
Structural priming paradigm also serves as a valuable tool for study of structural linguistic behaviour in LMs
~\citep{marvin-linzen-2018-targeted,van-schijndel-linzen-2018-neural,sinclair2022structural,michaelov-etal-2023-structural,jumelet-etal-2024-language}.
Parallels to human effects can be made with the in-context learning (ICL) behaviour of LMs, indeed, lexical repetition within a target has been found to boost and predict priming strength~\citep{sinclair2022structural,jumelet-etal-2024-language}. 
\citet{sinha-etal-2023-language} observe that the presence of syntactic content in the surrounding text can inadvertently prime the model thus biasing its evaluations -- an effect amplified in longer contexts.
LMs are also, if not more so than in humans~\citep{digutsch2023overlap}, susceptible to priming effects that are purely lexico-semantic. 
\citet{misra2020exploring} show that certain relationships between the words in the context systematically affect alterations in predictions.
When it comes to direct parallels to the extent to which LM priming predicts human priming, LMs predict human priming behaviour less well in the presence of shared function words between prime and target~\citep{SINCLAIR2026104713} - potentially indicating that they are more affected by word repetition than we are. 
While it is clear that context shapes priming in \textit{comprehension}, the extent to which this affects \textit{production} remains less clear.
Local word and word-sequence repetition patterns (often used as a marker of lexical priming behaviour) in LMs 
suggest that human-like lexical priming effects can be observed in production~\citep{molnar2023attribution}.
This is closely related to the investigation and explanation of ICL behaviours in LMs. The context from which a continuation is generated affects other properties beyond coherence. Indeed, at the same time, repetitions can reinforce co-occurrence even at the token level~\citep{yanunderstanding2024}, these patterns can be exploited to achieve greater performance, e.g. in reasoning tasks \cite{agarwal2024many}, and lead to undesirable repetitions in generated text \cite{li2023repetition}.
It remains an open question the extent to which structural priming can influence this, and whether repetition is led by lexical sequence matching, higher-level hierarchical repetition patterns, or indeed a combination of the two~\citep[e.g.][]{ahuja2024learning}.\looseness-1

% \vspace{-1em}
\section{Methods}
\label{sec:mehtods}
To investigate the effects of structural priming in language model generation, we conduct controlled experiments using a carefully controlled dataset \Cref{sec:data}, and sentence completion paradigm \Cref{sec:sentence_completion}.

\subsection{Data - PrimeLMDative+}
\label{sec:data}
We construct a hybrid dataset \textit{PrimeLMDative+} combining \textbf{dative} (\textsc{PO}, \textsc{DO}) sentences from PrimeLM corpus \cite{sinclair2022structural} and \textbf{non-dative} structures from BLiMP corpus \cite{warstadt2020blimp}
\footnote{
        Information on specific sub-corpora used is found in~\Cref{app:dataset}.
    }, 
PrimeLM CORE-transitive corpus, and sentences from ROC story cloze dataset \citep{mostafazadeh2016corpus} which we hand-annotate from $\sim$300 stories to ensure there are no dative structures. We sample sentences from them such that we extract a balanced set of sentences of \textsc{po}, \textsc{do} and \textsc{non-dative} structures. 
To enrich dataset with lexical diversity, and to ensure similar vocabulary across conditions, syntactic variation, and increased sentence length, we apply controlled data augmentation techniques, resulting in a dataset of over one million sentences (1,048,576). Specifically, we augment four levels: vocabulary, temporal complement, indirect speech, and negation\footnote{Details of how augmentation was applied, and a full list of augmentation materials used can be found in~\Cref{app:dataset}.~\Cref{tab:augmentation_examples} shows examples of augmentations.}. 

\begin{table}
\centering
\small
\setlength{\tabcolsep}{6pt}
\renewcommand{\arraystretch}{1.15}
\begin{tabular}{p{0.185\linewidth}p{0.70\linewidth}}
\toprule
\textbf{Condition} & \textbf{Example prompt} \\
\midrule
 & \textit{prime} \hfil \textit{target stem} \\
\textit{No context} & \hfill The student sent the \\

\textit{Non-dative} & Roger is lucky. \hfill The student sent the \\

\textit{Unrelated}  & ...gave a toy to a man. \hfill The student sent the \\

\textit{Coherence}  & ...gave a toy to a \textbf{man}. \hfill The \textbf{man} sent the \\
\bottomrule
\end{tabular}
\caption{
Sentence-completion examples. 
In the \textit{No context} condition, the target stem with a dative verb is presented without a preceding prime. 
For \textit{Non-dative} the prime is any non-dative construction, for \textit{Unrelated}, the prime and target are semantically independent, with no lexical overlap and $<0.5$ semantic similarity, for \textit{Coherence} the target reuses last animate noun from the prime, creating local discourse coherence.
}
\label{tab:prompt_condition_examples}
\end{table}
     
\subsection{Sentence Completion Experiment}
\label{sec:sentence_completion}
The core of our experiments is to test LM generation behaviour when completing a \textit{target} sentence stem given a \textit{prime} sentence in the preceding context, under varying conditions.
We thus construct prime and target stem pairs---as in examples (1) and (2)---for the three context structures (\textbf{PO, DO, \textit{\textbf{non-dative}}}). In addition, we vary the relationship between the prime and the target stem to assess how lexico-semantically related context modulates the incidence of priming (see~\Cref{tab:prompt_condition_examples}).

\paragraph{No Context} We include a setting where there is no priming sentence as a baseline setting to observe likelihood of each potential completion given a dative target stem. In this context, we expect the lowest incidence of dative generations with respect to settings where we prime the models. In addition, we expect LMs to generate more PO than DO constructions, in line with previous findings that PO structures are more frequent in English corpora, and are generally preferred over DO in contexts of syntactic ambiguity \citep{bresnanpredicting,Buckle}; and documented LM biases towards PO structures~\citep{jumelet-etal-2024-language}. We also expect that overall structural preferences may reflect frequency effects stemming from models' training data \citep[e.g.][]{Gries2004}.

\paragraph{Unrelated} This setting ensures that there is minimal lexico-semantic relatedness between the prime and the target stem. In line with findings from ~\citet{sinclair2022structural}, we expect priming effects to be weak in this condition.
    For PrimeLM dative sentences (PO and DO), we use the CORE prime--target pairings from the original dataset, which meet these criteria. For BLiMP, ROC and transitive PrimeLM primes (\textit{non-dative}), we pair them with target stems containing dative verbs from PrimeLM, ensuring minimal lexical overlap and low semantic similarity between the verbs.\footnote{
        We ensure no overlapping words, and exclude all pairs where the cosine similarity between prime and target verbs, computed using spaCy's \textsc{en\_core\_web\_md}, exceeds 0.5.
    }

\paragraph{Coherent} This category ensures lexical coherence between the prime and the target by reusing the most recent animate noun from the prime in the target stem (see example 2b). We construct this setting for PO, DO, \textit{non-dative} by replacing the noun in each \textit{Unrelated} target with that extracted noun
    \footnote{
        This process is straightforward for PrimeLM sentences; for ROC and BLiMP sentences that lack an animate noun, we use the gender-neutral subject pronoun \textit{They} to simulate narrative coherence.
        }.
    This setting mirrors the noun overlap condition in PrimeLM and reflects narrative coherence. In line with previous results in comprehension~\citep{sinclair2022structural}, we expect priming effects to be relatively stronger in this setting.

\subsection{LM sentence completion}
\label{sec:lmsentence_completion}
We evaluate four open-weight base model families: \textsc{Phi}, \textsc{Gemma}, \textsc{EuroLLM}, and \textsc{Qwen}, which differ in scale, training data composition, and multilingual coverage. This allows us to assess whether priming effects are robust across models with different pre-training profiles, while avoiding the confounds introduced by instruction tuning. Whenever possible, we compare smaller and larger models from the same family: Gemma-2-2B and Gemma-2-9B \citep{gemmateam2024gemma2improvingopen}, EuroLLM-1.7B and EuroLLM-9B \citep{martins2024eurollmmultilinguallanguagemodels}, and Qwen3-1.7B-Base and Qwen3-8B-Base \citep{qwen3technicalreport}. We also include Phi-2-2.7B \citep{li2023phi2} \citep{phi2} as a compact English-oriented baseline, since no directly comparable larger base model from the same generation is available in our experimental setting.
We generate sentence completions by giving each model a prompt consisting of a prime sentence (excluding the \textit{No Context} condition) followed by the start of a target sentence (the \textit{stem}). For each generated completion we retain only the first complete sentence.
This procedure is applied across all conditions and models.\footnote{Full implementation details, including generation parameters, decoding settings, and sentence extraction heuristics, are provided in~\Cref{app:generation_details} and~\Cref{app:sentencesplitting}, as well as a table of illustrative examples in ~\Cref{tab:illustrative_completions}.}

\subsection{Measuring Completion Alignment}
We investigate both the semantic and lexical alignment of the completion to the context. Sentence-level semantic similarity is measured with cosine similarity using \texttt{all-MiniLM-L6-v2} and \texttt{all-mpnet-base-v2}. Token-level semantic alignment is measured with BERTScore~F1. Surface-level lexical reuse is measured with Jaccard overlap and BLEU to capture both word- and n-gram level repetition. See \Cref{app:completion_alignment} for further details.

%----------------------------------------------
\section{Annotating Dative Structures}
\label{sec:classifier}
We annotate the LM completions as containing a Prepositional Object (PO), Double Object (DO), or \textit{non-dative} structure. Given the scale of the generated data, we develop a custom BERT-based classifier to automatically label completions, in addition to manually annotating a small subset to confirm our results and validate our automatic approach. 

\subsection{Hand-annotation and agreement}
\label{human_annotated_analysis}
We manually annotate a sample of 2700 sentences: 100 PO, 100 DO, and 100 \textit{non-dative} generations for each experimental condition (\textit{no prime, unrelated, coherence}), for each model generation set. 
Annotation was completed by authors, in consultation with linguist colleagues for more ambiguous cases.
Inter-annotator agreement, measured by Cohen's Kappa, was substantial to almost perfect, (Kappa = 0.92 for Phi-2-2.7B, 0.71 for Gemma-2-2B, 0.89 for EuroLLM-1.7B).
Through our hand-annotation of model-generated sentence completions (results of which are reported in~\Cref{app:human-annotated-additional}), we obtain a reliable ground-truth set of sentences against which we can evaluate our classification model,  as well as compare to our scaled-up automatic annotation findings. 
This hand-annotation is also valuable for flagging up potentially ambiguous cases, which are either far less likely in one alternation (3), or form part of an embedded clause (4). More examples are reported in~\Cref{appendix_annotations_details}.

{
\small

\ex. The brother showed the pilot \textbf{how to} navigate the dense forest.\hfill \emph{non-dative}

\ex. The pilot saved the plane from crashing, \textbf{and} then returned the iron to the student.\hfill \emph{non-dative}

}

\begin{table*}[ht!]
    \centering
    \small
    \scalebox{0.8}{
    \begin{tabular}{llcccccccccc}  
        \toprule
        \multirow{2}{*}{\textbf{Model}} 
        & \textbf{Test Set} 
        & \textbf{Accuracy} 
        & \multicolumn{3}{c}{\textbf{Precision}} 
        & \multicolumn{3}{c}{\textbf{Recall}} 
        & \multicolumn{3}{c}{\textbf{F1-score}} \\
        \cmidrule(lr){4-6} 
        \cmidrule(lr){7-9} 
        \cmidrule(lr){10-12}
        & & 
        & \textbf{PO} & \textbf{DO} & \textbf{Non-dative} 
        & \textbf{PO} & \textbf{DO} & \textbf{Non-dative} 
        & \textbf{PO} & \textbf{DO} & \textbf{Non-dative} \\
        \midrule

        \multirow{4}{*}{Logistic Regression} 
        & PrimeLMDative+      & 0.96 & \textbf{0.99} & 0.98 & 0.91 & \textbf{1.00} & 0.88 & 0.98 & \textbf{1.00} & 0.93 & 0.95 \\
        & Gemma-2-2B   & 0.56 & 0.67 & 0.20 & \textbf{0.85} & \underline{0.67} & \underline{0.67} & 0.50 & \textbf{0.67} & 0.30 & 0.63 \\
        & EuroLLM-1.7B  & 0.59 & 0.56 & 0.09 & \textbf{0.72} & 0.48 & 0.17 & \textbf{0.69} & 0.51 & 0.12 & \textbf{0.71} \\
        & Phi-2-2.7B          & 0.52 & 0.43 & 0.21 & \textbf{0.92} & 0.70 & \textbf{0.79} & 0.45 & 0.53 & 0.34 & \textbf{0.60} \\
        \midrule

        \multirow{4}{*}{BERT} 
        & PrimeLMDative+      & \textbf{1.00} & 1.00 & 1.00 & 1.00 & 1.00 & 1.00 & 1.00 & 1.00 & 1.00 & 1.00 \\
        & Gemma-2-2B   & \textbf{0.90} & 0.72 & 0.83 & \textbf{0.93} & 0.73 & 0.59 & \textbf{0.95} & 0.72 & 0.69 & \textbf{0.94} \\
        & EuroLLM-1.7B  & \textbf{0.86} & 0.50 & 0.85 & \textbf{0.90} & 0.54 & 0.58 & 0.92 & 0.52 & 0.69 & 0.91 \\
        & Phi-2-2.7B          & \textbf{0.89} & 0.67 & \textbf{0.94} & 0.92 & 0.75 & 0.65 & 0.94 & 0.71 & 0.77 & 0.93 \\
        \bottomrule
    \end{tabular}
    }
    \caption{Classifier comparison on the in-domain PrimeLM+ test set and on hand-annotated generated completions from different language models. We compare the final BERT classifier against a linguistically motivated Logistic Regression baseline using hand-engineered syntactic features. Accuracy, precision, recall, and F1-scores are reported for each class (PO, DO, and Non-dative). BERT shows more robust performance across generated test sets and is therefore used for structural labelling in the main experiments. We provide additional details in~\Cref{app:classifier_selection}.}
\label{tab:classifier_results}
\end{table*}

\subsection{Automatic annotation}

\paragraph{Approach} We choose to fine-tune a BERT-based classification model \citep{devlin-etal-2019-bert} to distinguish between Prepositional Object (PO), Double Object (DO), or \textit{non-dative} when given an input sentence. 
\citet{yao-todd-2024-berts} have shown that embeddings from pre-trained BERT models can successfully predict dative alternations in naturalistic corpora, distinguishing between PO and DO forms with good accuracy. 
Complementary evidence from child dialogue data further indicates that BERT-based approaches are effective for classifying dative structures in noisier, less well-formed contexts \citep{liu-wulff-2023-development}---since our sentences are relatively well-formed, we expect this approach to be effective for our purposes. 
We extract all unique sentences of PO, DO and \textit{non-dative} structures from \textit{PrimeLMDative+}~(\Cref{sec:data}), ensuring an even distribution across the three labels.\footnote{To prevent data leakage, we first partition the dataset into training and test sets, and only then apply data augmentation separately to each split, following the procedure described in~\Cref{app:augmentation}.}
We use an 80-20 train-test split, and standard 
finetuning hyperparameters for classification \footnote{Full BERT classifier details can be found in~\Cref{app:bert_details}.}
We also explore whether a Logistic Regression classifier that makes use of linguistically motivated features including POS tags and dependency parses of the sentences would prove a comparable approach, but find that while it proves to be a strong baseline, the Bert-based model is more stronger (Table~\ref{tab:classifier_results}).\footnote{Further Logistic Regression details can be found in ~\Cref{app:logistic_regression_details}}

\paragraph{Performance}
Finally, to validate our automatic classification pipeline, we test our BERT-based classifier on two different test corpora: a) the in-domain testset of \textit{PrimeLMDative+}, b) the LM generated sentences that we hand annotate, comparing the BERT-based labels with human annotations. The agreement between the human labels and the BERT classifier was consistently high, with an accuracy of 0.89 for Phi-2-2.7B, 0.90 for Gemma-2-2B, and 0.86 for EuroLLM-1.7B\footnote{Full results are reported in~\Cref{tab:classifier_results}.}. 
Example completions can be primed ($P$) \textit{The prince saved the computer for the secretary. A professor gave a \underline{paper to a student.} (PO, Phi)}, or non-primed ($\overline{P}$) \textit{A winner threw a juice to a husband . A husband kept the \underline{juice and drank it .} (PO, EuroLLm)}
Additional analyses on the hand-annotated samples, including preliminary structural priming patterns, semantic similarity, and lexical overlap, are reported in~\Cref{app:human-annotated-additional}.

\begin{figure*}[tb]
    \centering
    \begin{subfigure}[t]{0.48\textwidth}
        \centering
        \includegraphics[width=\textwidth]{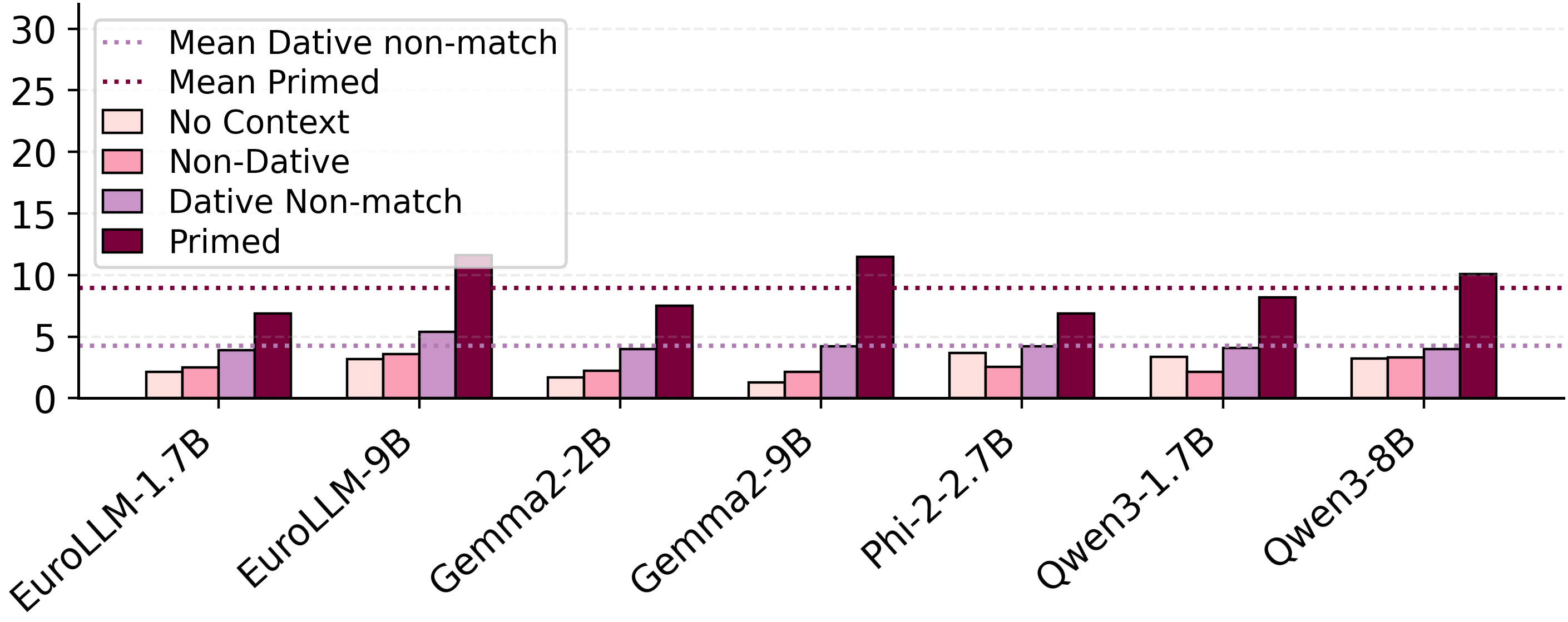}
        \caption{DO}
    \end{subfigure}%
    ~
    \begin{subfigure}[t]{0.48\textwidth}
        \centering
        \includegraphics[width=\textwidth]{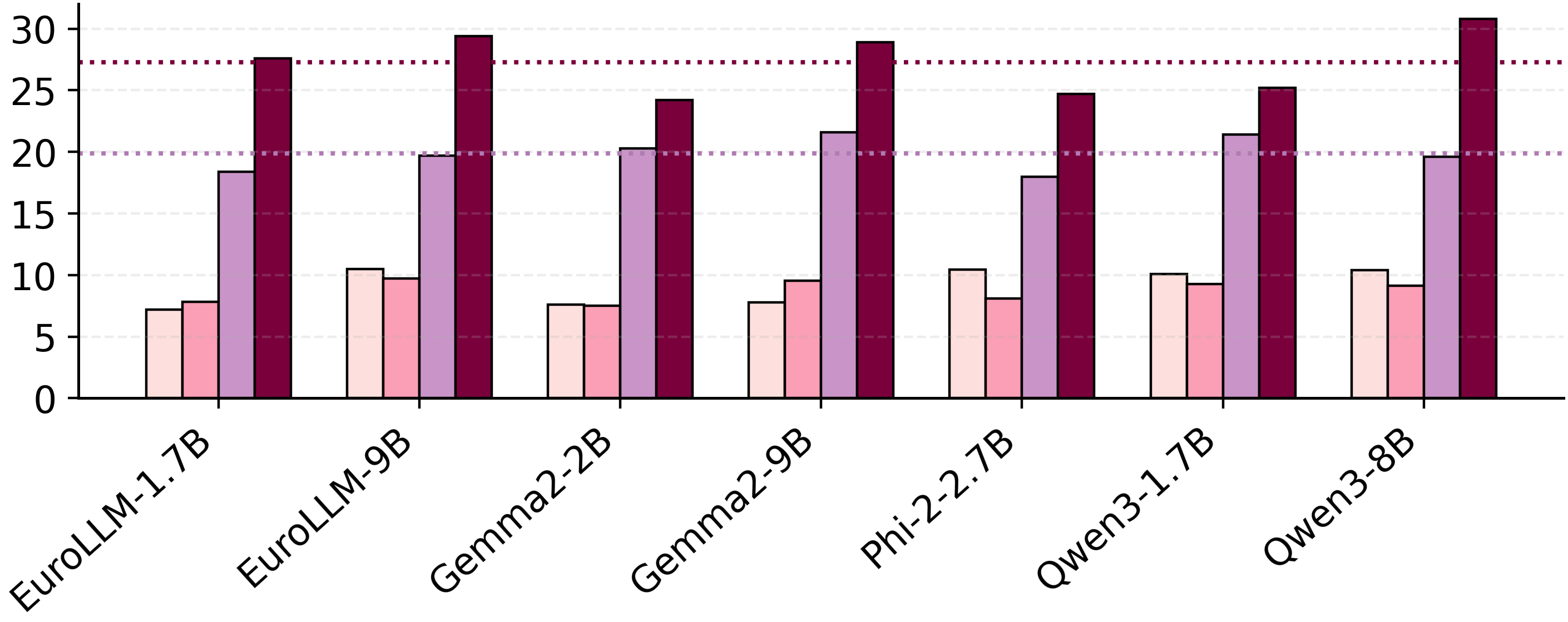}
        \caption{PO}
    \end{subfigure}
    \caption{Structure-matching completion rates for DO (left) and PO (right) structures under four conditions: \textit{No Context}, \textit{Non-dative}, \textit{Dative Non-match}, and \textit{Primed}. The \textit{Dative Non-match} condition contains a dative prime whose structure does not match the target structure. Dotted horizontal lines indicate the average rate across models for the \textit{Dative Mon-match} and \textit{Primed} conditions. Across both target structures, the matching primed condition produces the highest rates of structure-matching completions, indicating structural priming beyond both non-dative and non-matching dative controls. Full results are reported in~\Cref{app:human-annotated-additional},~\Cref{tab:rq1_priming_three_baselines}.}
    \label{fig:rq1}
\end{figure*}

% \input{sections/3_methods.tex}
% \input{sections/4_results_priming}
%----------------------------------------------
\section{Results: Priming in Production}
\label{sec:results}

We now investigate the extent to which LMs exhibit structural priming in production, and how this is modulated by lexico-semantic relationships within context and completions.
We first test whether models exhibit structural priming independently of semantic coherence, by comparing structure-matching completions in the \textit{unrelated} condition against a \textit{no context} baseline and a \textit{non-dative} baseline (\Cref{subsec:rq1_baselines}). We then examine whether priming strength varies as a function of contextual factors, including semantic coherence and lexical repetition (\Cref{subsec:rq2_coherence_effects}). Finally, we assess whether structural priming is accompanied by lexico-semantic alignment, measuring whether primed completions show increased semantic similarity and lexical overlap with their primes (\Cref{subsec:rq3_semantic_evidence_full}).

\subsection{Structural priming}
\label{subsec:rq1_baselines}
\paragraph{Presence of priming}
To confirm the presence of structural priming (RQ1), we test whether a given prime increases the probability of producing a matching structure relative to three baselines. The first of which, \textsc{no context}, captures each model's tendency to produce \textsc{po} or \textsc{do} completions from the target stem alone. The second consists of a \textsc{non-dative} sentence, which captures each model's tendency to produce dative completions given a prior, neutral sentence. 
The third, and hardest baseline against which to confirm greater incidence of structurally primed completions, is \textsc{dative non-match}, where the incidence of a structure given a prime of the \textit{alternative} structure is measured, e.g., a \textsc{PO} completion given a \textsc{DO} prime. In this final setting, a model completes a sentence containing a dative verb given a preceding dative sentence.
To test the presence of priming, we analyse only the stimuli where the prime and target sentence are not semantically coherent (the \textit{unrelated} condition), so that any increase in structure-matching completions can be attributed to \textit{structural} priming rather than semantic facilitation.

For each model $m$ and target structure $s \in \{\textsc{PO}, \textsc{DO}\}$, we define two priming effects:
% \[
$\Delta^{N}_{m,s} =
P_m(\hat{y}=s \mid U_s) - P_m(\hat{y}=s \mid N),
% \]
% \[
\Delta^{O}_{m,s} =
P_m(\hat{y}=s \mid U_s) - P_m(\hat{y}=s \mid O)$,
% \]
where $U_s$ denotes the \textit{unrelated} condition with prime structure $s$, $N$ denotes the \textit{no context} baseline, and $O$ denotes the non-dative prime baseline. Positive values indicate that the model more frequently produces structure-matching datives following a prime than in an unprimed baseline.

We compute distribution of generated structures conditioned on prime structure. For PO-prime items, we measure the percentage of completions classified as \textsc{PO}; for DO-prime items, we measure the percentage classified as \textsc{DO}. We then compare these rates against both baseline conditions. This allows us to test whether observed increase in dative production is driven by exposure to a matching dative frame, rather than by the target stem alone or by the mere presence of preceding context.
\begin{figure*}[tb]
    \centering
    \begin{subfigure}[t]{0.48\textwidth}
        \centering
        \includegraphics[width=\textwidth]{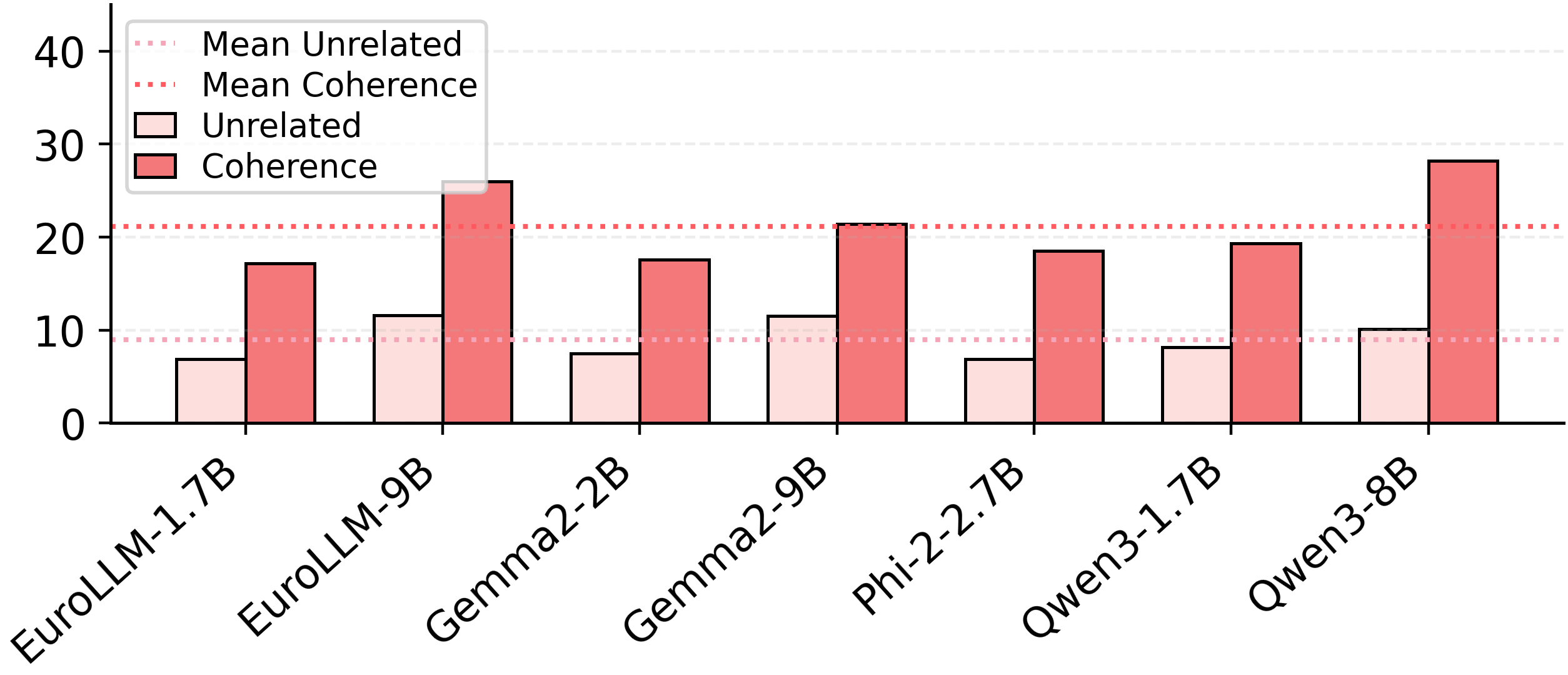}
        \caption{DO}
    \end{subfigure}%
    ~
    \begin{subfigure}[t]{0.48\textwidth}
        \centering
        \includegraphics[width=\textwidth]{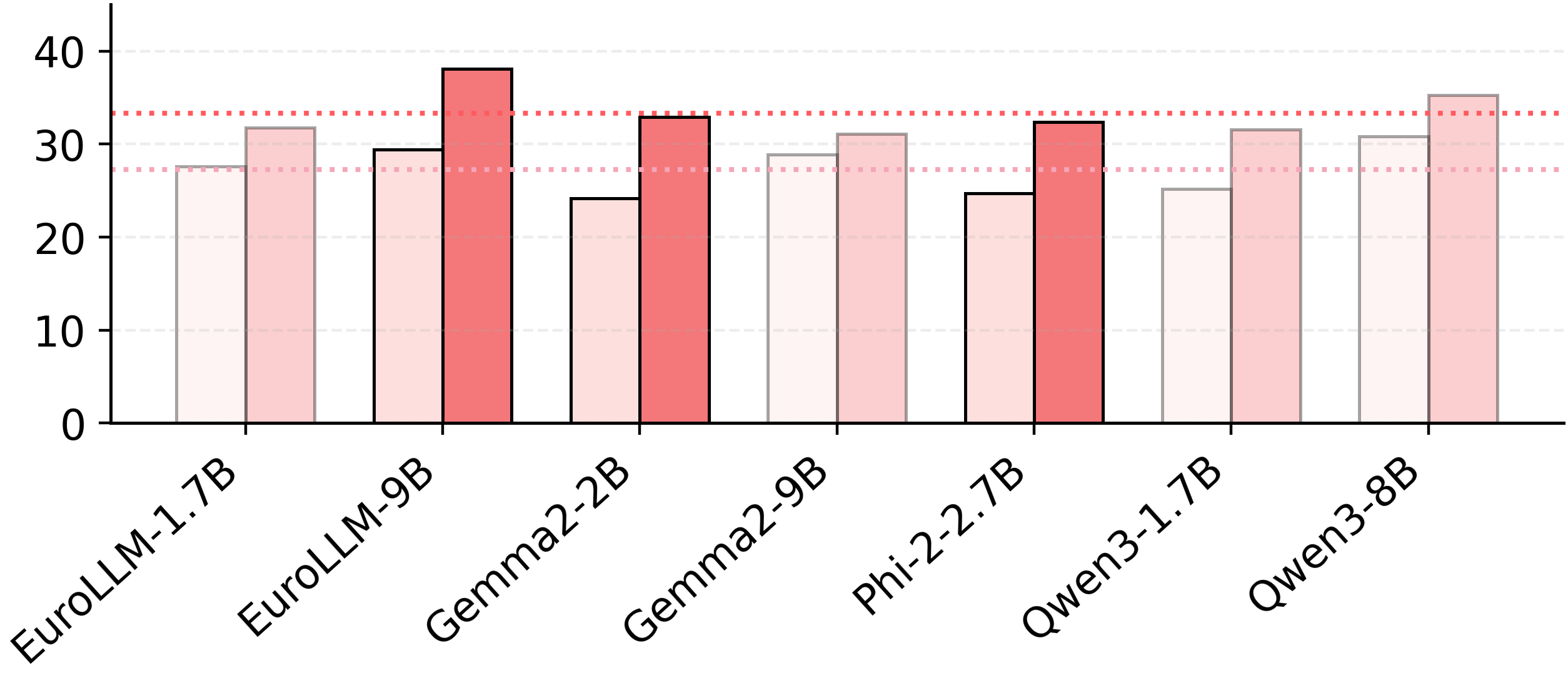}
        \caption{PO}
    \end{subfigure}
    \caption{Structure-matching completion rates for DO (left) and PO (right) targets in the \textit{unrelated} and \textit{coherence} conditions. For both target structures, the coherence condition tends to produce higher rates of structure-matching completions than the unrelated condition, showing that semantic coherence amplifies structural priming. Lighter bars mark PO comparisons for which the coherence--unrelated difference is not significant after Benjamini--Hochberg FDR correction. Full results are reported in~\Cref{app:human-annotated-additional},~\Cref{tab:rq2-coherence-unrelated}.}
    \label{fig:rq2}
\end{figure*}

Tests are implemented with \textit{proportions\_ztest}. The binary outcome is whether the generated continuation matches $s$. For a baseline $B \in \{N,O\}$, the test statistic is:
% \[
$z^{B}_{m,s} =
\frac{\hat{p}_{U_s} - \hat{p}_{B}}
{\sqrt{\hat{p}(1-\hat{p})
\left(n_{U_s}^{-1} + n_{B}^{-1}\right)}}$,
% \]
where $\hat{p}_{U_s}$ and $\hat{p}_{B}$ are the observed proportions in the \textit{unrelated} dative-prime condition and the baseline condition, respectively, and $\hat{p}$ is the pooled proportion. We correct model-level comparisons using Benjamini--Hochberg FDR correction and treat corrected $p < .05$ as significant.
As shown in~\Cref{fig:rq1}, for all models, and for both target structures, the \textit{unrelated} dative-prime condition produces higher rates of structure-matching completions than both the \textit{no-context} baseline and the \textit{non-dative} baseline. This indicates that exposure to a matching dative frame increases the likelihood of reproducing that structure even in the absence of semantic coherence, and thus provides strong evidence of structural priming in LM production.\footnote{
We additionally confirm that these priming effects are not solely due to POS sequence repetition, and indeed are more hierarchical in nature (similar to the structural complexity analysis in~\citet{sinclair2022structural}), finding the majority of our completions ($\sim$70\% (63\% DO, 74\% P0) are longer and thus more complex than the prime. Details supplied in~\Cref{app:length_analysis}, and a full breakdown in ~\Cref{app:stat_tests_for_main}~\Cref{tab:rq1_priming_three_baselines}.}

\paragraph{Relative priming strength across structures}
We observe that the absolute difference in proportion of primed completion versus all three of the baselines is higher for \textit{PO} rather than \textit{DO} structures (see the horizontal lines in Figure~\ref{fig:rq1} for the average rate in \textit{Primed} vs the \textit{Dative Non-match} baseline).
This is contradictory to our expectations of a greater priming effect in DO, the less expected construction, given results in comprehension~\citep[e.g.][]{sinclair2022structural,jumelet-etal-2024-language}.

\begin{figure}
    \centering
    \includegraphics[width=\linewidth]{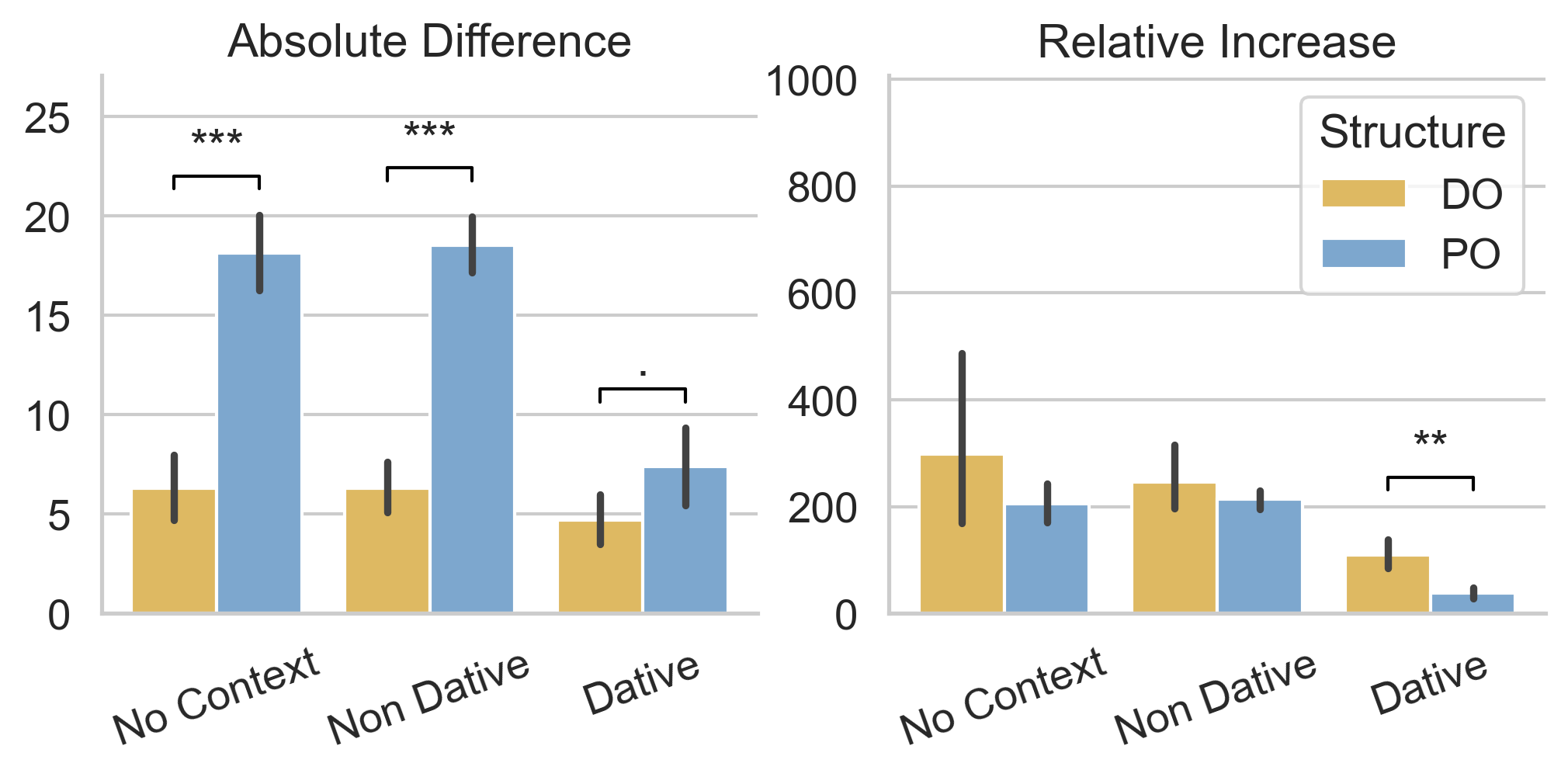}
    \caption{Priming strength relative to \textit{No context, Non dative} and \textit{Dative} (non-matching) baselines measured by i) absolute difference and ii) relative increase in production rate of PO and DO constructions. Results averaged across LMs, * indicates significant differences for t-tests comparing PO and DO priming strength. 
    }
    \label{fig:rq1a}
\end{figure}

To examine this further, we calculate i) the absolute difference in proportion, and ii) the percentage increase in proportional incidence between primed structure rate and each baseline of the datives we consider. 
As can be observed in Figure~\ref{fig:rq1a}, we find that as well as producing a greater number of POs overall, primed LMs produce POs with a greater absolute increase than DOs against our baselines, with borderline significance for the \textsc{Dative Non-matching}. 
Primed DO constructions meanwhile, exhibit the greater \textit{relative} increase in incidence, with significantly higher relative increase against the strongest baseline. This greater relative increase is in line with inverse frequency effects observed in comprehension, suggesting that while inverse frequency plays a role in production priming, these effects are potentially attenuated by the greater entropy associated with producing DO constructions, or hidden by the larger number of POs produced.\footnote{Extended results reported in~\Cref{app:priming_strength}}

\begin{table*}[tb]
\centering
\scriptsize
\setlength{\tabcolsep}{3.2pt}
\resizebox{\textwidth}{!}{
\begin{tabular}{ll|rrr|rrr|rrr|rrr|rrr|rrr|rrr}
\toprule
\textbf{Metric} & \textbf{Prim.}
& \multicolumn{3}{c|}{\textbf{Phi-2}}
& \multicolumn{3}{c|}{\textbf{Gemma-2-2B}}
& \multicolumn{3}{c|}{\textbf{Gemma-2-9B}}
& \multicolumn{3}{c|}{\textbf{EuroLLM-1.7B}}
& \multicolumn{3}{c|}{\textbf{EuroLLM-9B}}
& \multicolumn{3}{c|}{\textbf{Qwen-3-1.7B}}
& \multicolumn{3}{c}{\textbf{Qwen-3-8B}} \\
\cmidrule(lr){3-5}
\cmidrule(lr){6-8}
\cmidrule(lr){9-11}
\cmidrule(lr){12-14}
\cmidrule(lr){15-17}
\cmidrule(lr){18-20}
\cmidrule(lr){21-23}
& & \textbf{$\overline{P}$} & \textbf{$P$} & \textit{t}
  & \textbf{$\overline{P}$} & \textbf{$P$} & \textit{t}
  & \textbf{$\overline{P}$} & \textbf{$P$} & \textit{t}
  & \textbf{$\overline{P}$} & \textbf{$P$} & \textit{t}
  & \textbf{$\overline{P}$} & \textbf{$P$} & \textit{t}
  & \textbf{$\overline{P}$} & \textbf{$P$} & \textit{t}
  & \textbf{$\overline{P}$} & \textbf{$P$} & \textit{t} \\
\midrule

MiniLM & U
& 0.26 & 0.36 & 8.78
& 0.28 & 0.38 & 8.65
& 0.27 & 0.36 & 8.18
& 0.23 & 0.35 & 11.20
& 0.23 & 0.33 & 9.80
& 0.27 & 0.35 & 6.96
& 0.26 & 0.35 & 7.86 \\
& C
& 0.36 & 0.48 & 12.54
& 0.38 & 0.49 & 13.22
& 0.39 & 0.53 & 15.05
& 0.32 & 0.44 & 13.29
& 0.36 & 0.48 & 14.64
& 0.36 & 0.46 & 10.68
& 0.37 & 0.48 & 12.35 \\

\midrule

MPNet & U
& 0.26 & 0.35 & 8.40
& 0.27 & 0.37 & 8.52
& 0.26 & 0.34 & 7.76
& 0.23 & 0.34 & 11.03
& 0.23 & 0.32 & 9.41
& 0.26 & 0.34 & 6.53
& 0.26 & 0.33 & 6.83 \\
& C
& 0.34 & 0.44 & 11.67
& 0.36 & 0.46 & 12.05
& 0.37 & 0.49 & 13.52
& 0.31 & 0.42 & 12.56
& 0.34 & 0.44 & 12.80
& 0.35 & 0.43 & 9.68
& 0.35 & 0.45 & 11.09 \\

\midrule

BERTScore F1 & U
& 0.21 & 0.26 & 8.55
& 0.18 & 0.22 & 6.14
& 0.18 & 0.23 & 7.76
& 0.15 & 0.20 & 7.08
& 0.17 & 0.21 & 6.24
& 0.15 & 0.21 & 8.42
& 0.16 & 0.21 & 7.49 \\
& C
& 0.26 & 0.31 & 9.23
& 0.20 & 0.25 & 8.48
& 0.21 & 0.27 & 11.28
& 0.18 & 0.23 & 7.89
& 0.20 & 0.25 & 9.56
& 0.19 & 0.23 & 7.08
& 0.19 & 0.25 & 9.47 \\

\midrule

Jaccard & U
& 0.07 & 0.14 & 11.84
& 0.07 & 0.14 & 11.46
& 0.07 & 0.14 & 12.58
& 0.06 & 0.14 & 14.92
& 0.06 & 0.14 & 13.75
& 0.07 & 0.14 & 11.23
& 0.07 & 0.14 & 12.29 \\
& C
& 0.12 & 0.24 & 17.20
& 0.12 & 0.22 & 15.77
& 0.12 & 0.26 & 17.56
& 0.10 & 0.21 & 16.87
& 0.12 & 0.25 & 19.46
& 0.12 & 0.22 & 14.68
& 0.13 & 0.26 & 17.57 \\

\midrule

BLEU & U
& 4.27 & 6.66 & 9.29
& 4.36 & 6.24 & 7.66
& 4.42 & 7.31 & 9.76
& 4.01 & 6.72 & 10.24
& 4.35 & 7.38 & 9.54
& 4.49 & 6.70 & 8.29
& 4.57 & 7.12 & 9.44 \\
& C
& 5.05 & 10.09 & 13.48
& 5.02 & 9.37 & 12.82
& 5.35 & 11.80 & 14.00
& 4.65 & 8.83 & 12.85
& \lightgraytext{5.33} & \lightgraytext{11.27} & \lightgraytext{15.23}
& 5.26 & 9.89 & 11.89
& 5.48 & 12.23 & 15.49 \\
\bottomrule
\end{tabular}
}
\caption{RQ3 results. Mean similarity scores for non-primed ($\overline{P}$) and primed ($P$) completions in the \textit{unrelated} (U) and \textit{coherent} (C) conditions. Reported \textit{t}-values come from Welch’s \textit{t}-tests comparing $\overline{P}$ and $P$ completions within each context. The same results are shown in~\Cref{fig:rq3}. All comparisons are statistically significant after Benjamini--Hochberg FDR correction ($p<.001$ in all cases), except for the \lightgraytext{EuroLLM-9B in coherence}. In the Appendix we report the breaking down for PO (see~\Cref{tab:po_priming_similarity_tests}), and DO (see~\Cref{tab:do_priming_similarity_tests}). The results show a less consistent effect for DO in the \textit{Unrelated} condition.}
\label{tab:rq3}
\end{table*}

\subsection{Coherence Effects on Structural Priming}
\label{subsec:rq2_coherence_effects}

To address RQ2, we test whether semantic coherence between prime and target boosts structural priming. We compare \textit{coherence} and \textit{unrelated} conditions while holding prime structure fixed. This allows us to ask whether models are more likely to reproduce primed dative structure when the target stem is semantically linked to the preceding prime, rather than merely preceded by a structurally matching but semantically unrelated context.

For each model $m$ and target structure $s \in \{\textsc{PO}, \textsc{DO}\}$, we define the coherence effect as:
% \[
$\Delta^{C-U}_{m,s} =
P_m(\hat{y}=s \mid C_s) - P_m(\hat{y}=s \mid U_s)$,
% \]
where $C_s$ denotes the \textit{coherent} condition with prime structure $s$, and $U_s$ denotes the \textit{unrelated} condition with the same prime structure. Positive values of $\Delta^{C-U}_{m,s}$ indicate that coherence increases the probability of generating the primed structure.

We apply the same two-sample proportion \textit{z}-test in the \textit{coherent} and \textit{unrelated} conditions following~\S~\ref{subsec:rq1_baselines}. The results show that coherence generally increases structure-matching completions. For PO primes, the coherence advantage is positive for all models, although the effect is not statistically reliable for Gemma-2-9B after correction. For DO primes, the coherence effect is larger and consistent across all models, with coherent DO primes producing substantially more DO completions than unrelated DO primes. These results indicate that structural priming is not only present in semantically unrelated contexts, but is systematically amplified when the prime and target are semantically coherent. Full results are reported in~\Cref{fig:rq2} and~\Cref{tab:rq2-coherence-unrelated}.

\subsection{Lexico-semantic Alignment under Structural Priming}
\label{subsec:rq3_semantic_evidence_full}
To address RQ3, we test whether structural priming is accompanied by greater lexico-semantic alignment between the prime and the generated completion.
This allows us to examine whether structural repetition is associated with broader alignment effects during generation.

For each model and contextual condition, we split completions into structurally primed ($P$) and non-primed ($\overline{P}$) cases after being exposed to either a PO or DO structure. A completion is treated as $P$ when the generated structure matches the structure of the dative prime; otherwise, it is treated as $\overline{P}$. We then compare these two groups across the \textit{coherence} and \textit{unrelated} conditions.

We investigate both the semantic and lexical alignment of the completion to the context.  
For each metric $r$, model $m$, and condition 
$c \in \{\textit{coherence}, \textit{unrelated}\}$, we define the alignment effect of structural priming as the difference between the mean alignment score of structurally primed vs.
% completions and that of 
non-primed completions:
% \[
$\Delta^{r}_{m,c} =
\mathbb{E}\!\left[r \mid P, m, c\right]
-
\mathbb{E}\!\left[r \mid \overline{P}, m, c\right]$,
% \]
where $\mathbb{E}[\cdot]$ denotes the empirical mean over completions, $P$ denotes completions that realise the primed structure, and $\overline{P}$ denotes completions that do not. A positive value indicates that structurally primed completions align more strongly than non-primed completions.

We observe (see ~\Cref{tab:rq3}) a consistent increase in lexico-semantic alignment for structurally primed completions across models. This pattern is consistent across similarity metrics, and contextual conditions demonstrating that structural priming is not limited to syntactic repetition, and that both lexical and semantic repetition is boosted by structural priming. These effects are present in both PO and DO production individually, although stronger and more robust in PO completions. This stronger effect in PO may be related to their greater incidence, and facility of production, with repetition in DO more constrained.\footnote{A breakdown of results for PO and DO can be found in~\Cref{app:additional_analyses}, specifically Tables~\ref{tab:po_priming_similarity_tests}, \ref{tab:do_priming_similarity_tests}, and Figure~\ref{fig:rq3}.}
Effect sizes are consistently larger in \textit{coherent} rather than \textit{unrelated} primes. To investigate this further, we compare alignment between $P$ and $\overline{P}$ completions, finding comparable semantic alignment across coherent and unrelated conditions, but significantly greater \textit{lexical} repetition in primed completions in the coherent condition (see Figure~\ref{fig:alignment_lex_sim_coherent}, and~\Cref{app:alignment_rq3}), indicating lexical priming also plays a role.

\begin{figure}
    \centering
    \includegraphics[width=\linewidth]{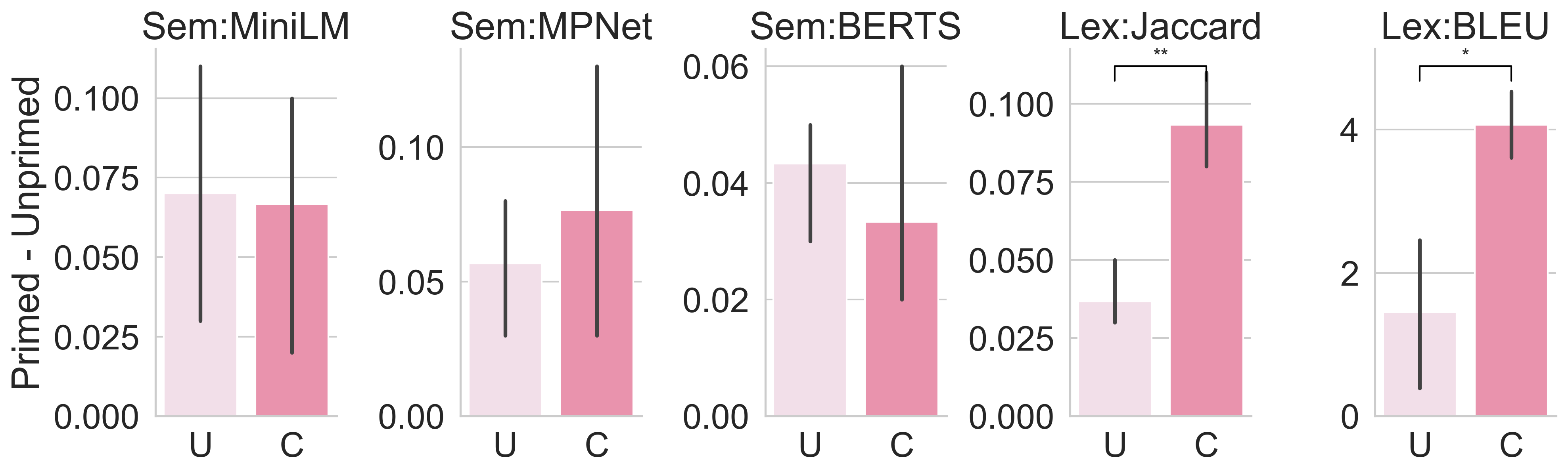}
    \caption{Lexico-semantic alignment boost in primed completions across \textbf{U}nrelated and \textbf{C}oherent settings. Results are an average across LMs. * Indicates independent t-tests comparing differences are significant.}
    \label{fig:alignment_lex_sim_coherent}
\end{figure}

\section{Discussion \& Conclusions}
Our results form an interesting piece of the puzzle when examining contextual factors that predict generation patterns in LMs.
In human structural priming, there is a degree of parity between comprehension and production~\citep{TooleyKM}, and since LM priming effects reported in a comprehension setting compare log probabilities between alternative constructions, we expect LM production to display similar properties.

We firstly observe that there is a greater incidence of POs produced by LMs than DOs, which we expect since DO is the less predictable structure~\citep{SINCLAIR2026104713}. In terms of the presence of priming (RQ1), when primed, LMs produce a higher level of the primed structure relative to multiple strict baselines, thus demonstrating a reliable priming effect. 
When comparing differences in priming strength across alternations, 
LMs exhibit a stronger \textit{relative} increase in incidence over a baseline for DOs than POs, in line with prior findings of LM priming, with a reliably stronger priming effect observed in DOs~\citep{sinclair2022structural,jumelet-etal-2024-language,SINCLAIR2026104713}. This is also in line with the inverse frequency effect in humans as noted by~\citep{jumelet-etal-2024-language,zhou-etal-2025-context}, whereby priming is higher in the less preferred structure.
However, we observe a stronger \textit{absolute} difference in primed incidence over the baseline for POs, the more frequently produced structure, indicating that in a production setting, these effects may be somewhat diluted by the greater preference for producing the more predictable PO structure.

In terms of lexico-semantic coherence boosting priming (RQ2), we observe that priming effects are reliably boosted by coherence between prime and target stem---in line with findings for humans~\citep{pickering1998representation, segaert2013syntactic, mahowald2016meta} and LMs~\citep{sinclair2022structural,jumelet-etal-2024-language}---and that this boost is greater and more reliable in DOs, the less predictable structure. 
Finally, in terms of structural priming promoting lexico-semantic alignment (RQ3), in line with lexical boost effects in humans~\citep{bock1986meaning, pickering1998representation, segaert2013syntactic, mahowald2016meta},
we observe that structurally primed completions contain significantly greater levels of lexico-semantic repetition than unprimed completions, which are comparable across conditions for semantic alignment, but are significantly higher for lexical repetition after coherent (and lexically overlapping) primes,
thus indicating that lexical priming is also at play in these settings.

The observed association between structural and lexico-semantic priming suggests that these mechanisms are closely linked and tend to co-occur, with greater lexical repetition associated with stronger structural priming, and vice versa. 
This relationship raises the question of whether the effects of repetition compound more strongly in LMs, which may be more sensitive to lexical repetition than humans~\citep{SINCLAIR2026104713}.

\paragraph{Outlook} Zooming out, our findings contribute to a body of evidence of human-like structural adaptation in LMs, expanding our understanding of parallels between LMs and humans for implicit learning, and expectation-based-processing.  
For example, \citet{Ziegler16032019} link structural priming in comprehension to information structure, and discourse expectations. Similarly, we find that coherence between prime and target stems strengthens priming, although further work is needed to disentangle this effect from a potential lexical boost.
Our work also opens questions concerning the mechanisms underlying how synergy between lexical and structural representations amplifies repetition. ~\citet{ICLR2024_yan_UnderstandingICL} conjecture that repetitions are a by-product of likelihood estimations influenced by the training data and find evidence of token co-occurrence reinforcement---that is, contextual co-occurrence strengthens their relationship. Our findings complement this work by finding evidence of structural-lexico-semantic priming: repetition effects are also compounded between words and the structures they occur in. 
Future work making use of interpretability techniques such as sequential attribution~\citep{sarti2023inseq} or context attribution~\citep{li2026attributing}---useful for understanding which tokens within the context drive repetitions~\citep{molnar2023attribution}---or Logit Lens~\citep{nostalgebraist2020interpreting}---which can identify \textit{where} within the layers an effect comes from~\citep{yan2026spurious}---can also clarify whether structural priming strengthens lexical repetition, or vice versa, and whether these effects are bidirectional or independent.

Overall, our results have implications for deeper understanding of ICL, and suggest that subtle combinations of lexical, semantic, and structural patterns in the input context may have a much larger influence on shaping the content in LM generations than previously thought~\citep{sinha-etal-2023-language, SINCLAIR2026104713}. This is important, for example when using LMs for simulating human production behaviour~\citep[e.g.,][]{utting-etal-2026-surprisal}, considering how the wording of queries can affect the safety of LM responses~\citep[e.g.,][]{pucci2026foodnoisefalse}, or when analysing populations of agentic LMs interacting for signs of cooperation~\citep[e.g.,][]{vallinder2025cultural}, or indeed, alignment.

\section*{Limitations}
One clear limitation of our findings, although extracted from a systematic targeted evaluation across a broad range of stimuli, is that our experiments only contain evidence in English for one alternating construction. Future work can extend our data and findings to other languages that contain alternative constructions, to examine if these production findings hold for other languages, or indeed cross-lingually.
Another limitation by nature of this study is there is no direct item-level comparison with human responses to the same stimuli, as noted in~\citep{SINCLAIR2026104713}, this can differ substantially at the item level in terms of effect directionality, thus in future work, we intend to gather human completions for the same stimuli, to enable a larger comparison of priming behaviour.

Finally, out of scope for this work, but very much of interest for exploring further influences of inverse frequency on LM behaviour is the impact of the lexical bias of the verbs in the dataset, as well as the impact of verb preference on the strength of the priming effects. As noted in \citet{jumelet-etal-2024-language}, verb preference is a predictor of priming strength, as is surprisal, and thus may help to untangle the relationship between priming strength and differences in production and comprehension behaviour. 

\section*{Acknowledgements}
We would like to gratefully acknowledge the members of The Context Lab for their helpful feedback on earlier versions of this work. We specifically thank Irina Giuliava and Elspeth Edelstein for their helpful annotations and expertise, and Mingrui Zhou for his support with access to the server.
We would also like to thank our metareviewer for their helpful comments, as well as express our gratitude for the helpful suggestions from the anonymous reviewers, which we have incorporated in our work. 

We disclose the use of generative AI tools for light editing and rephrasing; as well as with the refinement of plots and tables; the original text and the ideas within it are our own, and we carefully reviewed all suggested edits.

\bibliography{anthology,priming_production}

\appendix

\section{Dataset Creation Details}
\label{app:dataset}
This appendix expands the dataset construction procedure described in~\Cref{sec:data}. 
We provide additional details on the source corpora, sentence selection criteria, and augmentation steps used to build the final experimental dataset.

\paragraph{Dative sentences} We select only CORE conditions from the \textbf{PrimeLM} dataset as our dative prime-target pairs. 

\paragraph{Non-dative sentences} We select sentences from \textbf{Blimp}, where we specifically make use of 15 of their released 67 sub-datasets (animate subject passive, determiner noun agreement 1, determiner noun agreement 2, ellipsis n bar 1, ellipsis n bar 2, intransitive, passive 1, passive 2, regular plural subject verb agreement 1, regular plural subject verb agreement 2, sentential negotiation npi licensor present, superlative quantifiers, superlative quantifiers 2,tough vs raising 1, tough vs raising 2). 

We also annotate sentences from the \textbf{ROC} story cloze dataset, extracting 976 non-dative examples.

Finally, we use transitive sentences from the PrimeLM dataset. 

\paragraph{Augmentation}
To enrich the dataset with lexical diversity, syntactic variation, and increased sentence length, we apply controlled data augmentation techniques, resulting in a dataset of over one million sentences (specifically, 1,048,576). Augmentation was performed using a template-based approach, applied selectively to ensure coherence and grammaticality.
All three sources are augmented to share nouns in order that vocabulary is more diverse. We augment PrimeLM and ROC with indirect speech. We further augment PrimeLM sentences with all remaining augmentation strategies (vocabulary, temporal and negation).

\subsection{Augmentation Templates}
\label{app:augmentation}
~\Cref{tab:augmentation_templates} contains the materials with which we created the augmentations for our dataset. 

\begin{table*}[t]
    \centering
    \small
    \scalebox{0.9}{
    \begin{tabular}{lll}
        \toprule
        \textbf{Augmentation} & \textbf{PO Sentence} & \textbf{DO Sentence} \\
        \midrule
        \textbf{Original} & The teacher sent the book to the student. & She gave her mother a present. \\
        \textbf{Vocabulary} & Jasmine sent the book to the student. & Stacy gave her mother a present. \\
        % \hline
        \textbf{Temporal} & \textit{A couple of days ago,} the teacher sent the book to the student. & \textit{Last month,} she gave her mother a present. \\
        % \hline        
        \textbf{Indirect Speech} & \textit{Joana told Susan that} the teacher sent the book to the student. & \textit{Angela said that} she gave her mother a present. \\
        % \hline
        \textbf{Negation} & The teacher \textit{did not} send the book to the students. & She \textit{didn't} give her mother a present. \\
        \bottomrule
    \end{tabular}
    }
    \caption{Examples of PO and DO sentences and their augmentations (Vocabulary, Temporal, Indirect Speech, and Negation).}
    \label{tab:augmentation_examples}
    \vspace{-5pt}
\end{table*}

\paragraph{Vocabulary}
\label{app:augment_vocab}
To improve lexical variety and ensure an even distribution of vocabulary across syntactic classes, thereby reducing the risk of classifiers relying on superficial lexical heuristics~\citep[e.g.][]{ryb2022analog}, we compile a curated list of animate nouns drawn from our three source corpora. These are then used to generate augmented variants of each sentence within the sub-corpora.

\paragraph{Indirect Speech}
\label{app:augment_indirectspeech}
We further enrich the dataset by converting basic sentences into complex sentences using indirect speech constructions. This introduces an additional matrix clause with a reference verb (e.g., \textit{said}, \textit{noted}) followed by a complement clause introduced by \textit{that}, e.g., \textit{The reader said he found the gun for the club.} These additions increase syntactic depth and introduce distracting material, allowing us to test whether structural priming effects persist in more complex contexts.
Concretely, we design a template to create a reporting clause to be prepended to each original sentence in our dataset. We randomly select a subject from the noun list that we created, a verb that is allowed to introduce a reported clause, and the complementiser 'that'. In addition, when required by the verb, we add an indirect object from the same noun list, with the difference that instead of subject pronouns, we add indirect object pronouns.

We annotate the nouns in PrimeLM as to whether they have constraints to their likely personal pronouns e.g. (the woman: she/they, the lawyer: she/he/they) We make use of curated vocabulary lists of animate nouns from our corpus extracted by spaCy\footnote{\url{https://spacy.io/}} and annotated by hand.

\paragraph{Temporal Complement}
\label{app:augment_temporal}
To increase syntactic complexity at the surface level while preserving the target verb in the main clause, we either prepend or append temporal complements (e.g., \textit{last week}, \textit{yesterday}, \textit{a couple of days ago}) to PrimeLM sentences. 
The placement was not random, but respectful of our annotations regarding the possibility of adding each temporal complement at the beginning or at the end of sentences. In order to make the augmented sentences grammatically and semantically coherent, we also specify which temporal complement, when placed in a specific position in the sentence, requires a comma. We ensure that the temporal complements in our list are compatible with the past simple used in PrimeLM.
All templates are manually curated to ensure semantic and syntactic compatibility with the stimuli.

\paragraph{Negation}
Negative sentences introduce both structural alterations and semantic shifts, often requiring additional processing. 
Since BLiMP sentences often contain negated verbs by design, and ROC also contain negations and do not conform to a specific structure or tense, we only augment PrimeLM to include negations. 
We first identify the main verb and then apply rules to generate grammatically well-formed negative constructions using both formal (e.g., \textit{did not}) and informal (e.g., \textit{didn't}) negation markers. For example: \textit{She \underline{did not} buy him the book}, \textit{She \underline{didn't} buy him the book}.

\begin{table}[]
    \centering
    \begin{tabular}{p{0.45\textwidth}}
        \textbf{Augmentation Templates}\\
        \toprule
        \textit{Temporal Complement}\\
        \hline
         a couple of days ago, a few weeks ago, a few hours ago, a long time ago, at dawn, at dusk, early this morning, five minutes ago, last autumn, last summer, last spring, last winter, last week, last month, last year, yesterday.  \\
         \hline
         \textit{Indirect Speech verbs}\\
        \hline
         told, said, saw, heard, found out, read, wanted to say, wanted to tell, wanted to show, could have said, could have told, could have shown, should have said, should have told, should have shown \\
         \hline
         \textit{Vocabulary: ROC Names}\\
        \hline
         Abe, Allen, Allie, Andy, Angela, Anita, Ann, Anna, Annie, Arthur, Barry, Bea, Ben, Beverly, Bill, Billy, Bob, Bobby, Bonnie, Boston, Boy, Brad, Cam, Cara, Carl, Carly, Charles, Chris, Clint, Cora, Crystal, Dan, Daniel, David, Diana, Diane, Eliza, Ella, Ellie, Francine, Francis, Fred, Gary, Genie, Glenda, Greg, Hank, Heather, Jaime, Jake, James, Jane, Janet, Janie, Jasmin, Jasmine, Jason, Jean, Jeff, Jenn, Jenny, Jessica, Jill, Jo, Joan, Joanne, Jody, Joe, John, Johnnie, Joshua, Josie, Jude, Julie, July, June, Justin, Karl, Kaya, Kayla, Kelly, Ken, Kerry, Kira, Kyle, Lance, Larry, Laura, Leah, Lenny, Linda, Lisa, Lola, Lucy, Ludo, Luke, Marc, Marcus, Mario, Mark, Marlene, Martin, Mary, Matt, Meg, Megan, Melody, Meredith, Mia, Mike, Mikey, Naomi, Ned, Nick, Nicole, Niels, Noah, Oscar, Phil, Richard, Robbie, Roger, Rosie, Sam, Sammy, Sarah, Shane, Shannon, Shawn, Sonia, Sophia, Stacy, Stephanie, Stuart, Sue, Susan, Theo, Theodore, Tim, Timmy, Tom, Vanessa, Vera, Vivian, Walter, Wendy, \\
    \end{tabular}
    \caption{Augmentation Templates. Complements for Temporal augmentation, verbs for Indirect Speech augmentation, }
    \label{tab:augmentation_templates}
\end{table}

\section{Sentence Completion Generation Details}
\label{app:generation_details}
In this appendix we expand the sentence-completion procedure introduced in~\Cref{sec:lmsentence_completion}, providing the full generation parameters, and decoding choices.

\paragraph{Generation Settings}
To ensure consistency across models and conditions, we used a standardised generation procedure. 
All prompts were tokenised using each model's corresponding tokenizer. The end-of-sequence (EOS) token was used as the padding token, and left-padding was enabled to align batched input sequences. Inputs were padded and truncated to a maximum length of 512 tokens.

Sentence completions were generated under three predefined priming conditions: \textit{no-context}, \textit{unrelated}, and \textit{coherence}. For each prompt row, we generated one continuation using stochastic decoding with the following fixed parameters:
\begin{itemize}
    \item \textbf{Maximum generation length:} 40 new tokens.
    \item \textbf{Sampling:} \texttt{do\_sample=True}.
    \item \textbf{Top-$k$ filtering:} \texttt{top\_k=0}, corresponding to unrestricted top-$k$ sampling.
    \item \textbf{Batching:} Prompts were processed in batches of 16.
    \item \textbf{Inference mode:} \texttt{torch.no\_grad()} was used to disable gradient computation during generation.
\end{itemize}
Generated outputs were decoded with special tokens removed and saved together with the corresponding model identifier.
This standardised procedure was applied to all language models and experimental conditions, allowing us to isolate the effect of structural priming from other confounding variables.
We use a GeForce RTX 4090 with 24GB GPU RAM for all experiments.

\section{Sentence Splitting}
\label{app:sentencesplitting}
As mentioned in~\Cref{sec:lmsentence_completion}, in this appendix we provide full details of the post-processing steps, including sentence boundary detection and filtering.
After generation, to ensure that the syntactic labels assigned by BERT accurately reflected the structure of the first sentence generated by the language models, we developed a rule-based sentence extraction algorithm specifically designed to deal with structural irregularities commonly found in language model generations. This component was essential to isolate the intended completion and prevent misclassifications due to spill-over text or partial segmentation. To assess the reliability of the algorithm, we manually annotated a sample of 100 generated texts, in which the first sentence was carefully extracted by human annotators, becoming the gold-standard. We then applied our extraction algorithm to the same examples and conducted a strict string matching comparison against the gold annotations. The resulting accuracy was 98\%, confirming that the rule-based procedure is highly effective in identifying the relevant portion of text for subsequent classification, outperforming spacy's Sentencizer \cite{honnibal2020spacy} pipeline, as shown in~\Cref{tab:sentsplit_accuracy_comparison}.

Below, we describe each stage of the pipeline in detail, emphasising the fundamental importance of the order in which the various steps were executed.

We design a set of rules that result in a cumulative and dependency-aware algorithm:
\begin{itemize}
    \item Each step incrementally reduces ambiguity or protects structurally fragile elements;
    \item Subsequent steps build on the consistency and protections enforced by earlier operations;
    \item Later steps rely on the structural consistency and protection established earlier;
    \item The balance between precision (e.g., regex boundary detection) and recall (e.g., fallback to capitalised articles or post-processing) is maintained by carefully sequencing protective and corrective operations.
\end{itemize}
Reordering the steps would break this logic chain and substantially degrade the reliability of the extraction — as seen with the lower performance of baseline or hybrid methods.

\subsection{Text Cleaning}
To prepare the text for accurate sentence segmentation, we apply a series of cleaning operations to remove common artifacts of LM generation:
\begin{itemize}
    \item HTML entities and tags are removed to eliminate formatting noise;
    \item Line breaks are normalised by replacing newline characters with spaces;
    \item Sequences of whitespace characters are collapsed into a single space;
    \item Phrases such as ``Read more'' or ``Continue reading'' are removed when present, as they are often appended to generated texts and introduce noise in sentence-level analyses.
\end{itemize}

\subsection{Text Splitting Rules}

\begin{itemize}
    \item \textbf{Tokenisation} The input text is first tokenised to standardise spacing and character encoding.
    
    \item \textbf{Abbreviation and Decimal Protection} To prevent incorrect sentence splitting within common abbreviations (e.g., ``Dr.'') or numerical expressions (e.g., ``3.5''), full stops (.) are temporarily replaced with a protected token ([dot]) that preserves internal structure. This protection also extends to punctuation within brackets or quotation marks. After segmentation, all protected tokens are restored to their original form to preserve the intended phrasing.
    
    \item \textbf{Split Point Detection} We define a rule-based heuristic to identify the first sentence boundary. The primary strategy searches for punctuation marks (i.e., ``.'', ``!'', ``?'') followed by whitespace and an uppercase character, and selects the first such occurrence as the sentence boundary. If no such pattern is found, a secondary strategy searches for capitalised articles (e.g., ``The'', ``A'', ``An'') as potential fallback points. If neither heuristic applies, the entire string is retained.

    \item \textbf{Sentence Finalisation} If the resulting extracted segment does not end in terminal punctuation, a full stop (``.'') is appended to ensure sentence completeness.
\end{itemize}

\paragraph{Manual Evaluation}
To evaluate the reliability of the extraction pipeline, we manually annotated a sample of 100 examples by identifying the correct first generated sentence in each case. These gold first sentences were then compared to the output of the post-processing pipeline, resulting in an overall accuracy of 98\%.

\begin{table}[h!]
\centering
\begin{tabular}{|l|c|}
\hline
\textbf{Method} & \textbf{Accuracy} \\
\hline
spacy's Senticizer & 67\% \\
hybrid (cleaning + spacy) & 70\% \\
ours & \textbf{98\%} \\
\hline
\end{tabular}
\caption{Comparison of accuracy across sentence extraction methods.} 
\label{tab:sentsplit_accuracy_comparison}
\end{table}

\begin{table*}[t]
    \centering
    \small
    \setlength{\tabcolsep}{4pt}
    \renewcommand{\arraystretch}{1.15}
    \scalebox{0.9}{
    \begin{tabular}{p{1.7cm}p{1.8cm}p{4.0cm}p{2.8cm}p{3.2cm}p{1.2cm}}
        \toprule
        \textbf{Condition} 
        & \textbf{Prime Structure} 
        & \textbf{Prime sentence} 
        & \textbf{Target stem} 
        & \textbf{Completion} 
        & \textbf{Pred.} \\
        \midrule

        No Context 
        & -- 
        & -- 
        & The student sent a 
        & letter. 
        & non-dative \\

        \midrule

        \multirow{3}{*}{Unrelated}
        & Non-dative 
        & The princess opened the old book. 
        & The student sent a 
        & message to the professor. 
        & PO \\

        & \textbf{PO} 
        & The princess gave a \textbf{book} to the man. 
        & The student sent a 
        & \textbf{book} to the teacher. 
        & \textbf{PO} \\

        & DO 
        & The princess gave the man a book. 
        & The student sent a 
        & letter to his teacher. 
        & PO \\

        \midrule

        \multirow{3}{*}{Coherence}
        & Non-dative 
        & The princess opened the old book. 
        & They sent a 
        & letter to the prince. 
        & PO \\

        & \textbf{PO} 
        & The princess gave \textbf{a book} to the man. 
        & The man gave 
        & \textbf{a book} to a girl. 
        & \textbf{PO} \\

        & \textbf{DO} 
        & The princess gave the man a book. 
        & The man gave 
        & a girl \textbf{a book}. 
        & \textbf{DO} \\

        \bottomrule
    \end{tabular}
    }
    \caption{
    Illustrative model completions across prompt conditions. 
    Boldface in the \textbf{Prime Structure} and \textbf{Pred.} columns marks cases where the generated completion matches the structure of the prime, indicating structural priming. 
    Boldface within sentences highlights lexical material repeated between the prime and the generated completion. 
    The examples show that structurally primed completions can also contain greater lexical overlap with the prime, especially in coherent contexts.
    }
    \label{tab:illustrative_completions}
    \vspace{-5pt}
\end{table*}

\section{BERT Classification Details}
\label{app:bert_details}

We use a fine-tuned BERT-based sequence classifier to assign each generated completion to one of three structural classes: \textsc{PO}, \textsc{DO}, and \textsc{no context}. The classifier is implemented with the Hugging Face \texttt{transformers} library and loaded as an \texttt{AutoModelForSequenceClassification}. We use the label mapping \textsc{PO}~=~0, \textsc{DO}~=~1, and \textsc{no context}~=~2.

Input sentences are tokenised with the corresponding BERT tokenizer using truncation and padding, with a maximum sequence length of 512 tokens. At inference time, the model outputs class logits for each sentence, and the predicted class is obtained by taking the \texttt{argmax} over the logits. Predictions are generated under \texttt{torch.no\_grad()} to disable gradient computation.

The classifier was evaluated on the held-out portion of our multi-class dataset, identified by the \texttt{ttsplit = test} split. We report accuracy and class-level precision, recall, and F1-score using \texttt{sklearn.metrics}. The same prediction procedure is then applied to generated completions, where the predicted label is stored and used as the structural classification for the priming analyses.

During fine-tuning, training was performed for three epochs with a learning rate of \(2 \times 10^{-5}\), per-device training and evaluation batch sizes of 8, and weight decay set to 0.01. Evaluation was performed at the end of each epoch, with checkpoints saved during training. Mixed precision training was enabled to optimise GPU usage.

\subsection{Classifier selection: BERT vs. Logistic Regression}
\label{app:classifier_selection}

Before selecting the final classifier used for labelling generated completions, we compared a fine-tuned BERT classifier with a linguistically motivated Logistic Regression baseline. The Logistic Regression model was trained on hand-engineered syntactic features extracted with spaCy, including dependency relations relevant to dative alternation, the presence of prepositional markers such as \textit{to} and \textit{for}, POS-tag distributions, object-distance features, and whether the indirect object was realised as a pronoun or a noun phrase.

As shown in Table~\ref{tab:classifier_results}, Logistic Regression performs well on the in-domain PrimeLM+ test set, but its performance drops substantially on hand-annotated generated completions from several language models. In particular, it shows weaker and less stable performance on DO and non-dative cases across generated test sets. By contrast, BERT achieves consistently higher accuracy and more balanced class-level performance across both the in-domain test set and the model-generated evaluation sets. We therefore use the BERT classifier for the automatic structural labelling in the main experiments, while reporting Logistic Regression only as a transparent baseline.

\section{Logistic Regression Details}
\label{app:logistic_regression_details}

We train Logistic Regression on our multi-class dataset
% of sentences of \textsc{PO}, prepositional object; \textsc{DO}, double object; and \textsc{OTHER}, a selection of various structures
. We define custom linguistically motivated features for this classifier making use of dependency parse information extracted from SpaCy,
\footnote{
    \url{spacy.io} \textit{en\_core\_web\_sm}
}
 as well as other, structurally motivated features:

    \paragraph{{Object Distance} }
    The distance (in tokens) between the root verb and either the direct or prepositional object belonging to it, derived from the dependency parse of the sentence. This captures structural distinctions between PO and DO constructions: shorter distances typically correspond to DO structures.
    \paragraph{Specific Prepositions} 
    We include a feature that identifies and counts the presence of prepositional markers (e.g., \textit{to}, \textit{for}) appearing after the verb, to distinguish PO from DO structures, in which such prepositions are absent.
    
    \paragraph{Indirect Object Type}
    DO constructions frequently feature pronouns (e.g., \textit{him}, \textit{her}), whereas PO structures are more commonly associated with full noun phrases introduced by a preposition (e.g., \textit{to} the teacher). 
    This feature captures whether---if present---the indirect object in a sentence is a pronoun or a noun phrase.

    \paragraph{POS Tags} 
    We include part-of-speech (POS) tag features for the proportion of nouns, verbs, and prepositions present in a sentence. 
    These assist in differentiating between PO, DO, and non dative sentences.
    \paragraph{Dependency Parsing}
    We extract and quantify the frequency of structural relationships using the relations using the dependency parser from spaCy. Specifically, we use the presence of Subject-object dependencies (nsubj, dobj, iobj) and Prepositional attachments (prep, pobj) as our features.\footnote{We select these after investigating the relative importance for all relations, and extracting those which significantly differentiate between at least two classes}
    by computing the occurrence of specific dependency relations within each sentence, as identified by spaCy’s dependency parser. 
    These relationships include: Subject-object dependencies (nsubj, dobj, iobj): capturing whether a sentence contains a direct object, an indirect object, or both. Prepositional attachments (prep, pobj): identifying cases where an object is introduced by a preposition, which is a key feature distinguishing PO from DO constructions. 
    Auxiliary and modal verbs (aux, auxpass, modal): detecting auxiliary structures that may influence sentence complexity and argument structure. Nominal and adjectival dependencies (nmod, amod): capturing relationships that indicate noun phrase complexity, often relevant in distinguishing PO constructions.

\section{Completion Alignment Metrics}
\label{app:completion_alignment}

We compute completion-alignment metrics between the contextual prime and the generated continuation. 
Since the aim is to measure how much the generated completion aligns with the preceding context, all metrics are computed between the \texttt{prime\_sentence} and a cleaned completion-only field.

\paragraph{Completion extraction}
For each generated sentence, we first remove the target stem from the post-processed generation. 
More precisely, the field \texttt{completion\_only} is obtained by subtracting the \texttt{noun\_stub} string from \texttt{first\_generated\_PP\_clean} and stripping leading and trailing whitespace. 
Similarity metrics are then computed only for the \textit{unrelated} and \textit{coherence} conditions, since the \textit{no-context} condition does not contain a contextual prime against which the completion can be compared.

\paragraph{Sentence-level semantic similarity}
We compute sentence-level semantic similarity using two SentenceTransformer encoders: \texttt{all-MiniLM-L6-v2} and \texttt{all-mpnet-base-v2}. 
For each row, we encode the prime sentence and the completion independently and compute the cosine similarity between the two resulting sentence embeddings:
\[
\mathrm{sim}(p,c) =
\cos\bigl(\mathbf{e}(p), \mathbf{e}(c)\bigr),
\]
where \(p\) is the prime sentence, \(c\) is the generated completion, and \(\mathbf{e}(\cdot)\) is the sentence embedding produced by the corresponding encoder. 
The resulting scores are stored as \texttt{semsim\_all-MiniLM-L6-v2} and \texttt{sbert\_all-mpnet-base-v2}. 
Embeddings are computed in batches of 256.

\paragraph{Token-level semantic alignment}
We also compute BERTScore F1 between the completion and the prime sentence. 
The generated completion is treated as the candidate and the prime sentence as the reference. 
Following the implementation used in the notebook, we use \texttt{roberta-large} with English scoring, batch size 32, and baseline rescaling enabled:
\[
\mathrm{BERTScore}_{F1}(p,c).
\]
This metric captures token-level semantic correspondence while allowing for non-identical surface realisations.

\paragraph{Lexical overlap}
Surface-level lexical reuse is measured with Jaccard overlap. 
Both the prime and the completion are lowercased and tokenised using the regular expression \texttt{\textbackslash b\textbackslash w+\textbackslash b}. 
Jaccard overlap is then computed as:
\[
J(p,c) =
\frac{|T(p) \cap T(c)|}{|T(p) \cup T(c)|},
\]
where \(T(p)\) and \(T(c)\) are the token sets extracted from the prime and completion, respectively. 
If the union is empty, the score is set to zero. 

\paragraph{BLEU}
Finally, we compute sentence-level BLEU using \texttt{sacrebleu}. 
The completion is treated as the hypothesis and the prime sentence as the reference:
\[
\mathrm{BLEU}(p,c) =
\mathrm{sacreBLEU}(c, [p]).
\]
Unlike Jaccard overlap, which measures unigram set overlap, BLEU captures surface-level repetition at the level of ordered \(n\)-grams.

\section{Annotations Details}
\label{appendix_annotations_details}
During our manual annotation, a small number of completions were difficult to assign unambiguously to one of the three labels: PO, DO, or \textit{non-dative}. 
These cases typically involved verbs that can appear in dative-like configurations but whose generated continuation did not realise a genuine transfer event. 
For consistency, we label as PO or DO only completions that instantiate a dative alternation with a transferred theme and a recipient/beneficiary. 
Cases involving clausal complements, pronominal placeholders, or non-recipient prepositional phrases are therefore annotated as \textit{non-dative}. 
The examples below illustrate recurrent ambiguous cases.

{
\small

\ex. The business promised the mayor \textbf{it} would comply with all local laws regarding safety, operation and vehicles.\hfill \emph{\textit{non-dative}}

\ex. The manager promised the team \textbf{that} they would get bigger salaries.\hfill \emph{\textit{non-dative}}

\ex. A lady threw a veil \textbf{over} him. \hfill \emph{non-dative}

}

\section{Results: Additional Analyses}
\label{app:additional_analyses}

\subsection{Effect Sizes}
\label{app:effect_sizes}
Effect sizes are reported for Research questions 1 and 2 in Table~\ref{tab:cohens_h_summary}, and for RQ3 in Table~\ref{tab:cohens_d_summary}.

\begin{table*}
\resizebox{\textwidth}{!}{
\begin{tabular}{llccccccc}
\toprule
\textbf{Comparison} & \textbf{Structure} & \textbf{Phi-2} & \textbf{Gemma2-2B} & \textbf{Gemma2-9B} & \textbf{EuroLLM-1.7B} & \textbf{EuroLLM-9B} & \textbf{Qwen3-1.7B} & \textbf{Qwen3-8B} \\
\midrule
Unrelated vs.\ No Prime & PO & 0.38*** & 0.47*** & 0.57*** & 0.56*** & 0.49*** & 0.41*** & 0.52*** \\
Unrelated vs.\ No Prime & DO & 0.15*** & 0.29*** & 0.47*** & 0.24*** & 0.34*** & 0.21*** & 0.29*** \\
Unrelated vs.\ Non-Dative & PO & 0.46*** & 0.47*** & 0.51*** & 0.54*** & 0.51*** & 0.43*** & 0.56*** \\
Unrelated vs.\ Non-Dative & DO & 0.21*** & 0.26*** & 0.40*** & 0.21*** & 0.31*** & 0.29*** & 0.28*** \\
Unrelated vs.\ Dative Non-match & PO & 0.16*** & 0.09* & 0.17*** & 0.22*** & 0.23*** & 0.09* & 0.26*** \\
Unrelated vs.\ Dative Non-match & DO & 0.12** & 0.15** & 0.28*** & 0.13** & 0.23*** & 0.17*** & 0.24*** \\
Coherence vs.\ Unrelated & PO & 0.17*** & 0.19*** & 0.05 & 0.09* & 0.18*** & 0.14** & 0.10* \\
Coherence vs.\ Unrelated & DO & 0.36*** & 0.31*** & 0.27*** & 0.32*** & 0.38*** & 0.33*** & 0.47*** \\
\bottomrule
\end{tabular}}
\caption{\textbf{Cohen's $h$ }effect sizes for the RQ1/RQ2 two-proportion $z$-tests, across all seven models. Stars denote Benjamini--Hochberg FDR-corrected significance ($^{*}p<.05$, $^{**}p<.01$, $^{***}p<.001$).}
\label{tab:cohens_h_summary}
\end{table*}

\begin{table*}
\resizebox{\textwidth}{!}{
\begin{tabular}{llccccccc}
\toprule
\textbf{Metric} & \textbf{Context} & \textbf{Phi-2} & \textbf{Gemma2-2B} & \textbf{Gemma2-9B} & \textbf{EuroLLM-1.7B} & \textbf{EuroLLM-9B} & \textbf{Qwen3-1.7B} & \textbf{Qwen3-8B} \\
\midrule
MiniLM & Unrelated & 0.60*** & 0.55*** & 0.48*** & 0.70*** & 0.57*** & 0.44*** & 0.47*** \\
MiniLM & Coherence & 0.64*** & 0.65*** & 0.77*** & 0.69*** & 0.70*** & 0.55*** & 0.60*** \\
MPNet & Unrelated & 0.55*** & 0.53*** & 0.45*** & 0.68*** & 0.54*** & 0.41*** & 0.40*** \\
MPNet & Coherence & 0.59*** & 0.60*** & 0.68*** & 0.64*** & 0.61*** & 0.49*** & 0.53*** \\
BERTScore F1 & Unrelated & 0.53*** & 0.38*** & 0.44*** & 0.41*** & 0.34*** & 0.49*** & 0.43*** \\
BERTScore F1 & Coherence & 0.48*** & 0.44*** & 0.59*** & 0.40*** & 0.44*** & 0.36*** & 0.46*** \\
Jaccard & Unrelated & 0.91*** & 0.83*** & 0.87*** & 1.14*** & 1.01*** & 0.86*** & 0.87*** \\
Jaccard & Coherence & 1.05*** & 0.96*** & 1.12*** & 1.09*** & 1.09*** & 0.89*** & 0.98*** \\
BLEU & Unrelated & 0.86*** & 0.70*** & 0.86*** & 0.94*** & 0.88*** & 0.73*** & 0.80*** \\
BLEU & Coherence & 1.01*** & 0.96*** & 1.06*** & 0.99*** & 0.97*** & 0.89*** & 1.01*** \\
\bottomrule
\end{tabular}}
\caption{\textbf{Cohen's $d$} effect sizes for the RQ3 Welch $t$-tests (primed vs.\ non-primed completions) on lexico-semantic similarity metrics, across all seven models. Stars denote Benjamini--Hochberg FDR-corrected significance ($^{*}p<.05$, $^{**}p<.01$, $^{***}p<.001$).}
\label{tab:cohens_d_summary}
\end{table*}

\subsection{Priming Strength}
\label{app:priming_strength}

\begin{figure*}[t]
    \centering
    \begin{subfigure}[t]{0.48\textwidth}
        \centering
        \includegraphics[width=\linewidth]{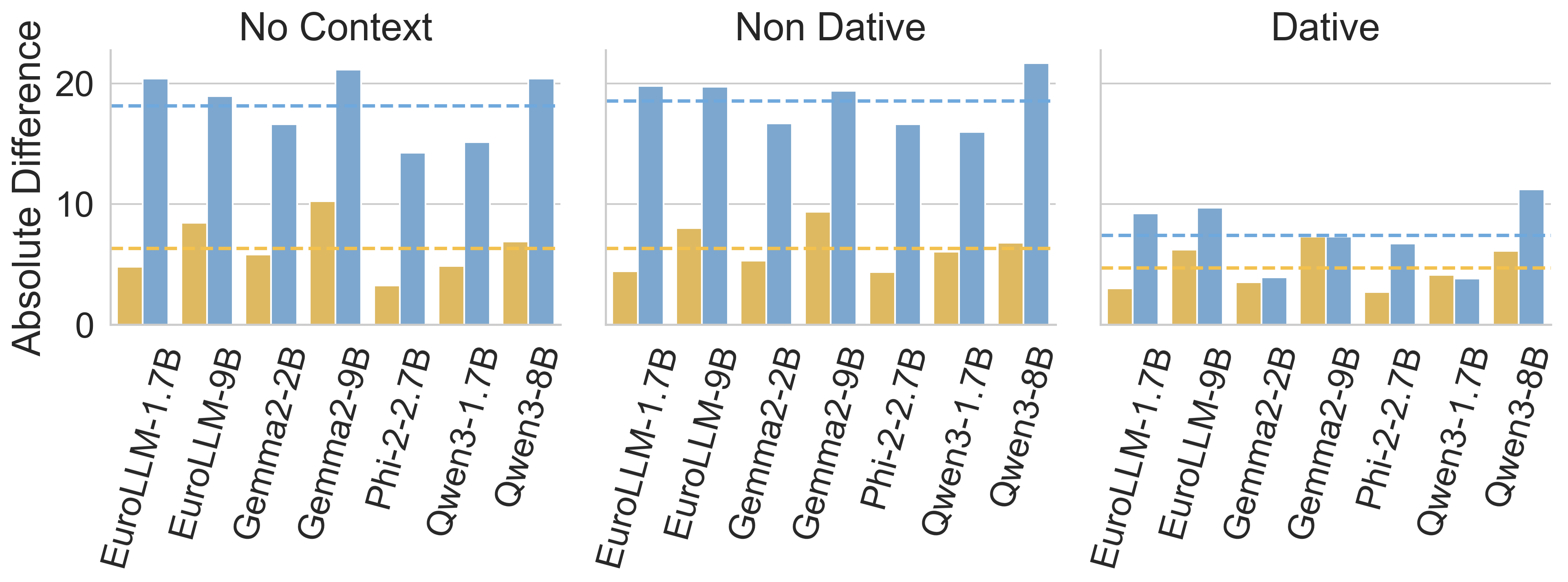}
        \caption{Absolute proportion difference}
        \label{fig:priming_strength_percent_difference}
    \end{subfigure}
    \hfill
    \begin{subfigure}[t]{0.48\textwidth}
        \centering
        \includegraphics[width=\linewidth]{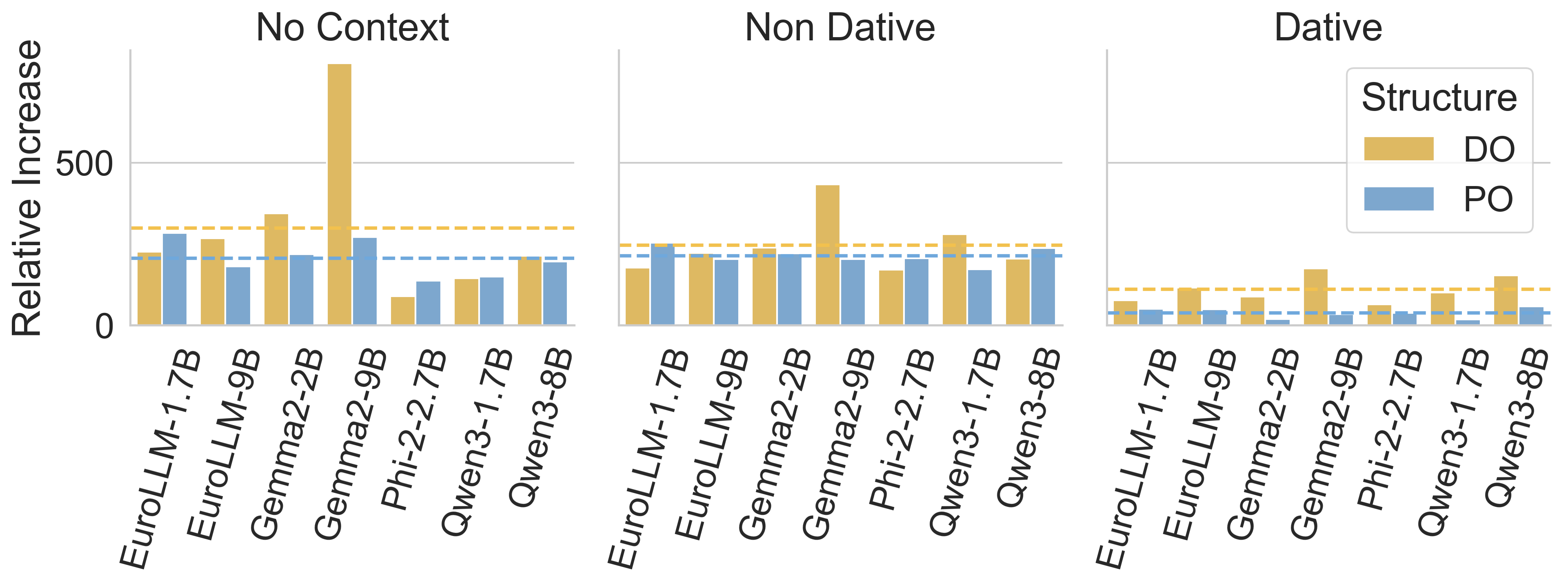}
        \caption{Relative proportion increase}
        \label{fig:priming_strength_percent_increase}
    \end{subfigure}
    \caption{\textbf{Priming strength} across models and baseline contexts, measured as (a) the absolute percentage-point difference and (b) the relative percentage increase.}
    \label{fig:priming_strength_appendix}
\end{figure*}

Priming strength, measured as the difference between primed incidence and baseline incidence, varies across models. Figure~\ref{fig:priming_strength_appendix} shows the per-model priming strengths for both i) Absolute difference in incidence and ii) Relative increase in proportion. t-tests for the average differences (as indicated by the dashed line, and reported in the main body of the text) can be found in Table~\ref{tab:priming_strength_tests}.

\begin{table}[h]
\centering
\small
\begin{tabular}{llcc}
\toprule
 & \textbf{Baseline} & \textbf{$t$} & \textbf{$p$} \\
\midrule
\multirow{3}{*}{\textit{Percent difference}}
& No Context & 8.516 & $< .0001$ \\
& Non-Dative & 11.420 & $< .0001$ \\
& Dative & 2.117 & .0599 \\
\midrule
\multirow{3}{*}{\textit{Percent increase}}
& No Context & -1.005 & .3500 \\
& Non-Dative & -0.919 & .3887 \\
& Dative & -4.440 & .0024 \\
\bottomrule
\end{tabular}
\caption{\textbf{Priming Strength:} Independent-samples $t$-tests comparing priming strength for PO and DO constructions under each baseline condition, using absolute percentage-point difference and relative percentage increase as effect measures.}
\label{tab:priming_strength_tests}
\end{table}

\subsection{Length Analysis}
\label{app:length_analysis}
To ensure that our priming results were not driven by structurally matching completions only, we analyse the lengths of the generated completions. We observe in Table~\ref{tab:priming_group_length_averages} that ~70\% of the primed completions are greater in length than the simple primes, assuring us that the effects found in the main paper are indicative of structural priming rather than simple POS sequence repetition.

\begin{table}[tb]
\centering
\small
\resizebox{\columnwidth}{!}{
\begin{tabular}{llccc}
\toprule
Group & Struct. & Mean Prime & Mean Generated & \% Same \\
\midrule
$\neg$Primed & DO & 7.00 & 11.93$^*$ & 5.42 \\
$\neg$Primed & PO & 8.00 & 12.60$^*$ & 8.75 \\
Primed & DO & 7.00 & 10.20$^*$ & 37.37 \\
Primed & PO & 8.00 & 12.40$^*$ & 26.17 \\
\bottomrule
\end{tabular}}
\caption{\textbf{Length Analysis}: Average prime length, generated-completion length, and percentage of same-length completions, by priming group. $^*$ denotes that the mean generated completion length is significantly larger than the average prime length with $p<.001$.}
\label{tab:priming_group_length_averages}
\end{table}

\subsection{Completion Alignment}
\label{app:alignment_rq3}
In addition to the statistical tests present in the main body of the text (Table~\ref{tab:rq3}), we also compare whether the boost in alignment varies by condition and thus is comparable between unrelated and coherent conditions.
Figure~\ref{fig:alignment_lex_sim_coherent} demonstrates that while the relative boost in semantic repetition is comparable, that lexical repetition is significantly higher in the coherent condition, demonstrating a lexical priming effect in our stimuli, since coherence was achieved in these cases via re-use of the last human Noun in the prime. Supporting t-test details are supplied in~\ref{tab:lex_sem_alignment}.

\begin{figure}
    \centering
    \includegraphics[width=\linewidth]{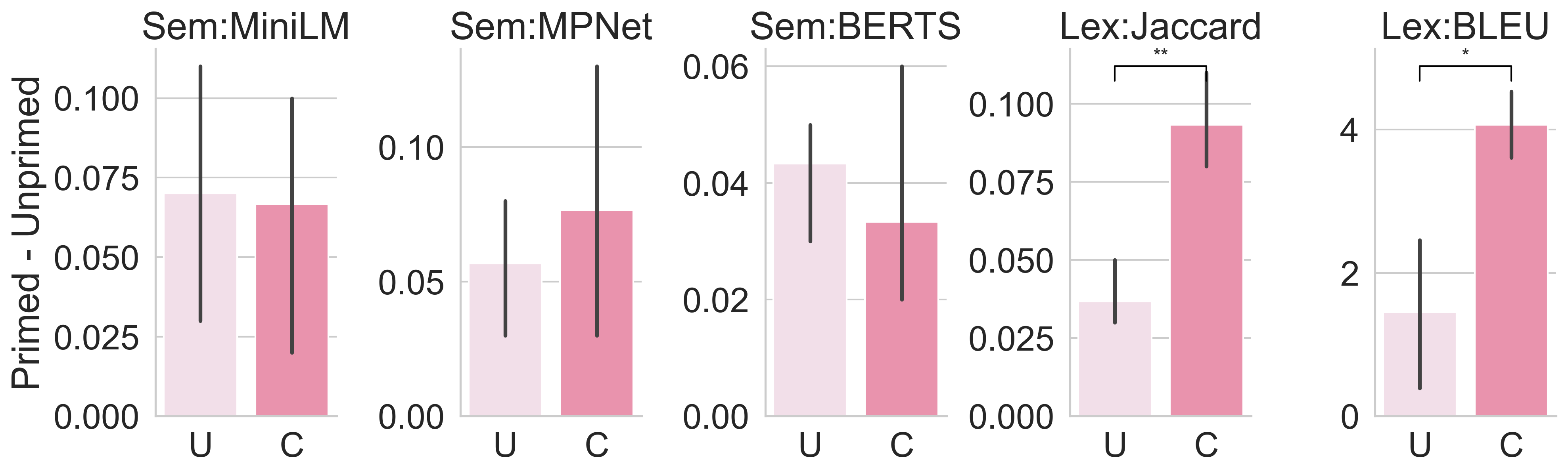}
    \caption{Lexico-semantic alignment boost in primed completions across \textbf{U}nrelated and \textbf{C}oherent settings. Results are an average across LMs. Independent t-tests compare differences, * indicates $p<0.05$.}
    \label{fig:alignment_lex_sim_coherent}
\end{figure}

\begin{table}[t]
\centering
\small
\begin{tabular}{lrrrr}
\toprule
\textbf{Metric} & \textbf{Mean U} & \textbf{Mean C} & \textbf{$t$} & \textbf{$p$} \\
\midrule
Jaccard      & 0.0367 & 0.0933 &  5.126 & .0083 \\
BLEU         & 1.4533 & 4.0700 &  3.998 & .0327 \\
BERTScore F1 & 0.0433 & 0.0333 & -0.671 & .5512 \\
MPNet        & 0.0567 & 0.0767 &  0.616 & .5825 \\
MiniLM       & 0.0700 & 0.0667 & -0.100 & .9252 \\
\bottomrule
\end{tabular}
\caption{\textbf{Lexico-semantic Alignment:} Comparison of lexical and semantic similarity between unrelated (U) and coherent (C) prime--target pairs. Comparisons performed grouping LM results.}
\label{tab:lex_sem_alignment}
\end{table}

\subsection{Supporting Statistics}
\label{app:stat_tests_for_main}
\begin{figure*}
    \centering
    \begin{subfigure}[t]{0.47\textwidth}
        \centering
        \includegraphics[width=\textwidth]{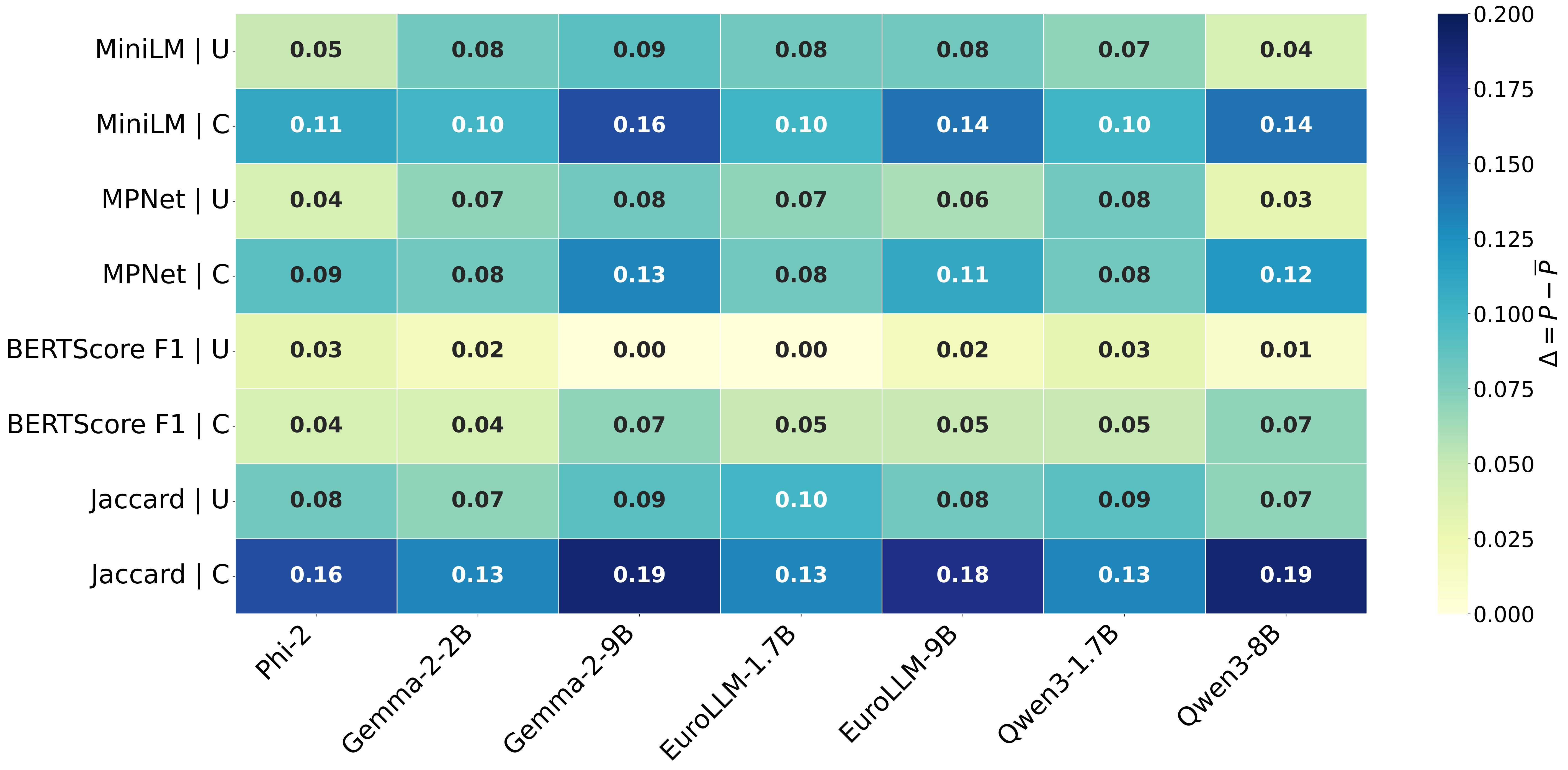}
        \caption{DO}
    \end{subfigure}%
    ~
    \begin{subfigure}[t]{0.47\textwidth}
        \centering
        \includegraphics[width=\textwidth]{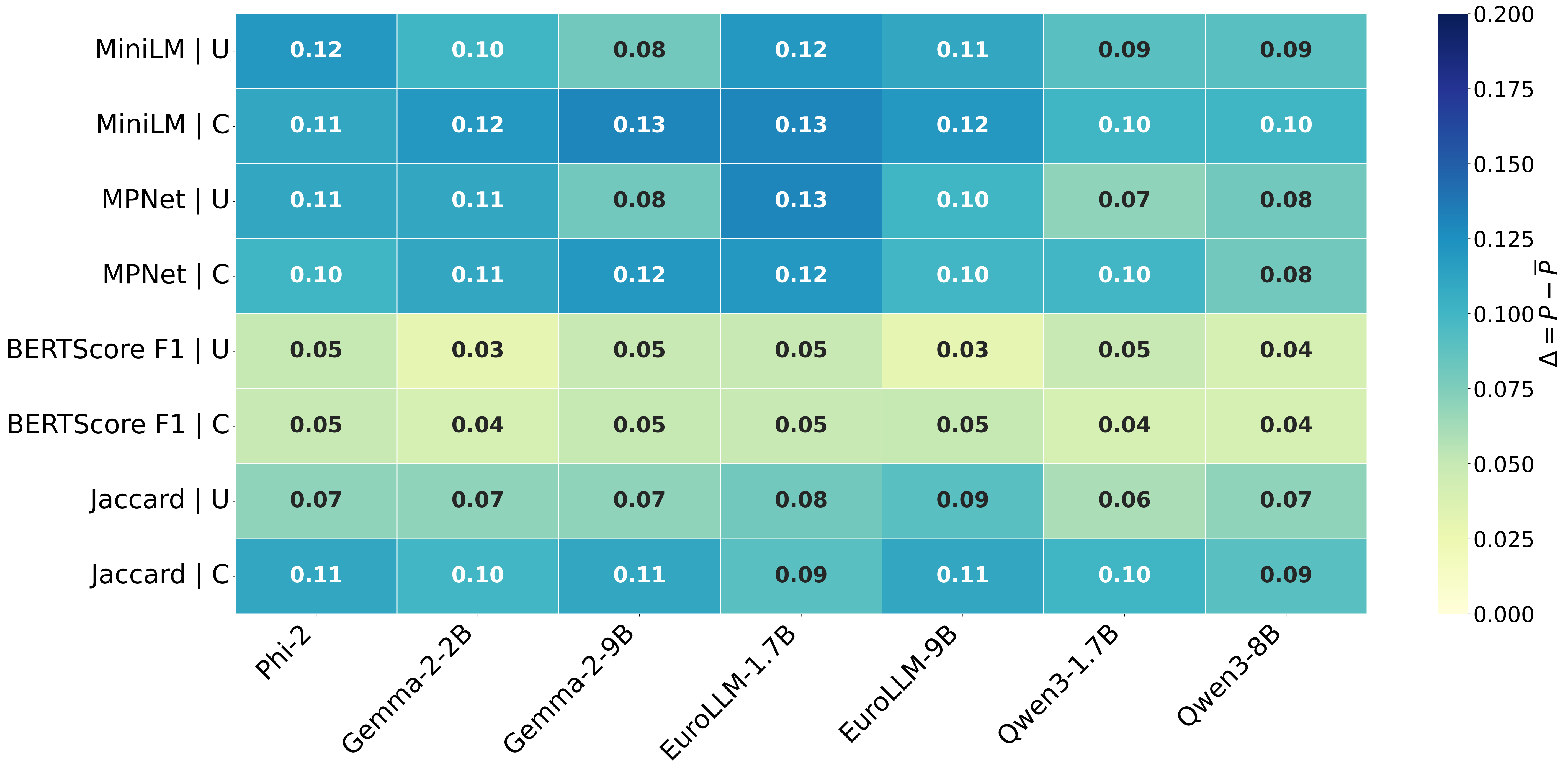}
        \caption{PO}
    \end{subfigure}
    \caption{Target-specific similarity analysis for PO and DO priming. Each heatmap reports $\Delta = P - \overline{P}$, where $P$ denotes completions that realise the primed structure and $\overline{P}$ denotes completions that do not. Rows correspond to similarity metrics under \textit{unrelated} (U) and \textit{coherence} (C) conditions; columns correspond to models. Positive values indicate that structurally primed completions are more similar to the prime than non-primed completions. The effect is consistently positive for PO and becomes stronger for DO in the coherence condition, suggesting that structural priming is accompanied by increased semantic and lexical alignment with the prime. We report the Welch's \textit{t}-tests results in~\Cref{tab:rq3}.}
    \label{fig:rq3}
\end{figure*}

To support the results reported in the main text concerning our three main research questions, RQ1 can be found here in Table~\ref{tab:rq1_priming_three_baselines}, RQ2 in Table~\ref{tab:rq2-coherence-unrelated}, with those for RQ3 reported in Table~\ref{tab:rq3} in the main paper, but with additional breakdowns reported here in Table~\ref{tab:po_priming_similarity_tests} and~\ref{tab:do_priming_similarity_tests}. The relative differences in lexico-semantic similarity across conditions is also presented in Figure~\ref{fig:rq3}.

\begin{table*}[t]
\centering
\scriptsize
\setlength{\tabcolsep}{3.2pt}
\renewcommand{\arraystretch}{1.08}
\begin{tabular}{llrrrrrrrrrr}
\toprule
\multirow{2}{*}{Model} 
& \multirow{2}{*}{Target}
& \multicolumn{3}{c}{\textbf{$\overline{P}$}}
& \multicolumn{1}{c}{\textbf{$P$}}
& \multicolumn{6}{c}{\textbf{Contrasts}} \\
\cmidrule(lr){3-5}
\cmidrule(lr){6-6}
\cmidrule(lr){7-12}
& 
& $\neg$N 
& $\neg$O 
& D
& P 
& $\Delta^{N}$ & $z^{N}$ 
& $\Delta^{O}$ & $z^{O}$ 
& $\Delta^{D}$ & $z^{D}$ \\
\midrule

\multirow{2}{*}{EuroLLM-1.7B} 
& PO & 7.20 & 7.81 & 18.40 & 27.60 & 20.40 & 16.37 & 19.79 & 14.11 & 9.20 & 4.89 \\
& DO & 2.12 & 2.49 & 3.90  & 6.90  & 4.78  & 7.04  & 4.41  & 5.66  & 3.00 & 2.97 \\
\midrule

\multirow{2}{*}{EuroLLM-9B} 
& PO & 10.47 & 9.70 & 19.70 & 29.40 & 18.93 & 13.97 & 19.70 & 13.40 & 9.70 & 5.04 \\
& DO & 3.16  & 3.60 & 5.40  & 11.60 & 8.44  & 9.95  & 8.00  & 8.26  & 6.20 & 4.97 \\
\midrule

\multirow{2}{*}{Gemma-2-2B} 
& PO & 7.59 & 7.53 & 20.30 & 24.20 & 16.61 & 13.64 & 16.67 & 12.38 & 3.90 & 2.10 \\
& DO & 1.69 & 2.22 & 4.00  & 7.50  & 5.81  & 8.72  & 5.28  & 6.76  & 3.50 & 3.36 \\
\midrule

\multirow{2}{*}{Gemma-2-9B} 
& PO & 7.78 & 9.53 & 21.60 & 28.90 & 21.12 & 16.53 & 19.37 & 13.26 & 7.30 & 3.76 \\
& DO & 1.27 & 2.16 & 4.20  & 11.50 & 10.23 & 13.84 & 9.34  & 10.40 & 7.30 & 6.07 \\
\midrule

\multirow{2}{*}{Phi-2-2.7B} 
& PO & 10.44 & 8.09 & 18.00 & 24.70 & 14.26 & 10.92 & 16.61 & 12.14 & 6.70 & 3.66 \\
& DO & 3.66  & 2.55 & 4.20  & 6.90  & 3.24  & 4.18  & 4.35  & 5.57  & 2.70 & 2.64 \\
\midrule

\multirow{2}{*}{Qwen-3-1.7B} 
& PO & 10.09 & 9.25 & 21.40 & 25.20 & 15.11 & 11.60 & 15.95 & 11.35 & 3.80 & 2.01 \\
& DO & 3.35  & 2.16 & 4.10  & 8.20  & 4.85  & 6.16  & 6.04  & 7.54  & 4.10 & 3.82 \\
\midrule

\multirow{2}{*}{Qwen-3-8B} 
& PO & 10.40 & 9.14 & 19.60 & 30.80 & 20.40 & 14.93 & 21.66 & 14.67 & 11.20 & 5.77 \\
& DO & 3.23  & 3.32 & 4.00  & 10.10 & 6.87  & 8.35  & 6.78  & 7.39  & 6.10  & 5.33 \\
\bottomrule
\end{tabular}
\caption{
RQ1. Presence of priming on the \textbf{full set} of generations: summary of  PO and DO completions after three $\overline{P}$ contexts, and a structure-matching prime. 
$\neg$ N denotes the no-context baseline; $\neg$ O denotes the non-dative prime baseline; D denotes the dative non-matching baseline, i.e., PO completions after DO primes or DO completions after PO primes. 
$P$ denotes the matching dative-prime condition, i.e., PO completions after PO primes or DO completions after DO primes. 
$\Delta^{N}$, $\Delta^{O}$, and $\Delta^{D}$ report the difference between $P$ and each baseline; $z^{N}$, $z^{O}$, and $z^{D}$ report the corresponding two-proportion z-test statistics. 
All percentage columns report structure-matching completions. All results are significant after Benjamini--Hochberg FDR correction at $\alpha=.05$.
This table shows full numerical results corresponding to~\Cref{fig:rq1}.}
\label{tab:rq1_priming_three_baselines}
\end{table*}

\begin{table*}[t]
\centering
\scriptsize
\setlength{\tabcolsep}{3.2pt}
\begin{tabular}{lrrrrc|rrrrc}
\toprule
& \multicolumn{5}{c|}{\% PO priming} & \multicolumn{5}{c}{\% DO priming} \\
\cmidrule(lr){2-6} \cmidrule(lr){7-11}
Model 
& Coher. & Unrel. & $\Delta^{C-U}$ & $z$ & $p_{\mathrm{BH}}$
& Coher. & Unrel. & $\Delta^{C-U}$ & $z$ & $p_{\mathrm{BH}}$ \\
\midrule
EuroLLM-1.7B & \lightgraytext{31.80} & \lightgraytext{27.60} & \lightgraytext{4.20} & \lightgraytext{2.06} & \lightgraytext{\hspace{0.8em}$.043$}
              & 17.20 & 6.90  & 10.30 & 7.07 & $<.001$ \\
EuroLLM-9B   & 38.10 & 29.40 & 8.70 & 4.11 & $<.001$
              & 26.00 & 11.60 & 14.40 & 8.24 & $<.001$ \\
Gemma-2-2B   & 32.90 & 24.20 & 8.70 & 4.31 & $<.001$
              & 17.60 & 7.50  & 10.10 & 6.82 & $<.001$ \\
Gemma-2-9B   & \lightgraytext{31.10} & \lightgraytext{28.90} & \lightgraytext{2.20} & \lightgraytext{1.07} & \lightgraytext{\hspace{0.8em} $.283$}
              & 21.40 & 11.50 & 9.90  & 5.97 & $<.001$ \\
Phi-2-2.7B   & 32.40 & 24.70 & 7.70 & 3.81 & $<.001$
              & 18.50 & 6.90  & 11.60 & 7.79 & $<.001$ \\
Qwen-3-1.7B  & \lightgraytext{31.60} & \lightgraytext{25.20} & \lightgraytext{6.40} & \lightgraytext{3.17} & \lightgraytext{\hspace{0.8em}$.002$}
              & 19.30 & 8.20  & 11.10 & 7.21 & $<.001$ \\
Qwen-3-8B    & \lightgraytext{35.30} & \lightgraytext{30.80} & \lightgraytext{4.50} & \lightgraytext{2.14} & \lightgraytext{\hspace{0.8em}$.038$}
              & 28.20 & 10.10 & 18.10 & 10.29 & $<.001$ \\
\bottomrule
\end{tabular}
\caption{RQ2 results comparing the \textit{coherence} and \textit{unrelated} conditions on the \textbf{full set} of generations.  
PO priming indicates PO incidence following a PO prime, DO computed analogously. 
$\Delta^{C-U}$ denotes the difference between \textit{coherent} and \textit{unrelated} percentages. Statistical reliability is assessed with two-sample proportion $z$-tests, with $p$-values corrected using Benjamini--Hochberg FDR correction. \lightgraytext{Gray text} indicates insignificant differences \textit{p $>$ 0.001}. This table shows full numerical results corresponding to~\Cref{fig:rq2}.}
\label{tab:rq2-coherence-unrelated}
\end{table*}

\begin{table*}[t]
\centering
\scriptsize
\setlength{\tabcolsep}{3.2pt}
\resizebox{\textwidth}{!}{
\begin{tabular}{ll|rrr|rrr|rrr|rrr|rrr|rrr|rrr}
\toprule
\textbf{Metric} & \textbf{Prim.}
& \multicolumn{3}{c|}{\textbf{Phi-2}}
& \multicolumn{3}{c|}{\textbf{Gemma-2-2B}}
& \multicolumn{3}{c|}{\textbf{Gemma-2-9B}}
& \multicolumn{3}{c|}{\textbf{EuroLLM-1.7B}}
& \multicolumn{3}{c|}{\textbf{EuroLLM-9B}}
& \multicolumn{3}{c|}{\textbf{Qwen-3-1.7B}}
& \multicolumn{3}{c}{\textbf{Qwen-3-8B}} \\
\cmidrule(lr){3-5}
\cmidrule(lr){6-8}
\cmidrule(lr){9-11}
\cmidrule(lr){12-14}
\cmidrule(lr){15-17}
\cmidrule(lr){18-20}
\cmidrule(lr){21-23}
& & \textbf{$\overline{P}$} & \textbf{$P$} & \textit{t}
  & \textbf{$\overline{P}$} & \textbf{$P$} & \textit{t}
  & \textbf{$\overline{P}$} & \textbf{$P$} & \textit{t}
  & \textbf{$\overline{P}$} & \textbf{$P$} & \textit{t}
  & \textbf{$\overline{P}$} & \textbf{$P$} & \textit{t}
  & \textbf{$\overline{P}$} & \textbf{$P$} & \textit{t}
  & \textbf{$\overline{P}$} & \textbf{$P$} & \textit{t} \\
\midrule

MiniLM & U
& 0.26 & 0.38 & 8.29
& 0.29 & 0.39 & 7.08
& 0.28 & 0.36 & 6.37
& 0.24 & 0.36 & 9.57
& 0.24 & 0.35 & 8.49
& 0.26 & 0.35 & 6.12
& 0.27 & 0.36 & 7.09 \\
& C
& 0.38 & 0.49 & 9.74
& 0.38 & 0.50 & 10.89
& 0.39 & 0.52 & 11.02
& 0.34 & 0.47 & 11.13
& 0.36 & 0.48 & 11.27
& 0.36 & 0.46 & 8.94
& 0.35 & 0.45 & 8.37 \\

\midrule

MPNet & U
& 0.25 & 0.36 & 8.38
& 0.27 & 0.38 & 7.67
& 0.27 & 0.35 & 6.13
& 0.23 & 0.36 & 9.75
& 0.23 & 0.33 & 8.44
& 0.26 & 0.33 & 5.54
& 0.27 & 0.35 & 6.41 \\
& C
& 0.36 & 0.46 & 8.90
& 0.36 & 0.47 & 10.38
& 0.36 & 0.48 & 10.20
& 0.32 & 0.44 & 11.37
& 0.34 & 0.44 & 9.73
& 0.34 & 0.44 & 8.45
& 0.33 & 0.41 & 7.85 \\

\midrule

BERTScore F1 & U
& 0.23 & 0.28 & 6.39
& 0.20 & 0.23 & 4.05
& 0.20 & 0.25 & 7.45
& 0.16 & 0.21 & 6.01
& 0.19 & 0.22 & 4.31
& 0.17 & 0.22 & 6.39
& 0.18 & 0.22 & 6.06 \\
& C
& 0.28 & 0.33 & 6.87
& 0.22 & 0.26 & 6.54
& 0.22 & 0.27 & 8.02
& 0.19 & 0.24 & 7.02
& 0.21 & 0.26 & 7.39
& 0.20 & 0.24 & 4.92
& 0.20 & 0.24 & 5.43 \\

\midrule

Jaccard & U
& 0.07 & 0.14 & 10.59
& 0.07 & 0.14 & 9.44
& 0.07 & 0.14 & 10.08
& 0.06 & 0.14 & 12.92
& 0.06 & 0.15 & 11.37
& 0.07 & 0.13 & 9.40
& 0.07 & 0.14 & 11.07 \\
& C
& 0.11 & 0.22 & 12.89
& 0.11 & 0.21 & 12.59
& 0.11 & 0.22 & 13.75
& 0.10 & 0.19 & 13.12
& 0.11 & 0.22 & 15.19
& 0.10 & 0.20 & 12.59
& 0.10 & 0.19 & 13.20 \\

\midrule

BLEU & U
& 3.95 & 6.41 & 9.50
& 4.04 & 5.94 & 7.76
& 3.99 & 6.96 & 9.40
& 3.78 & 6.43 & 11.46
& 3.95 & 7.05 & 9.20
& 4.12 & 6.27 & 8.42
& 4.15 & 6.76 & 8.86 \\
& C
& 4.17 & 7.71 & 11.30
& 4.48 & 7.60 & 11.59
& 4.47 & 7.69 & 10.23
& 4.22 & 7.41 & 10.99
& 4.49 & 7.60 & 12.80
& 4.39 & 7.48 & 11.22
& 4.36 & 7.24 & 10.01 \\

\bottomrule
\end{tabular}
}
\caption{RQ3. PO-specific similarity analysis on the \textbf{full set} of generations. Mean similarity scores for completions that do not realise the primed PO structure ($\overline{P}$) and completions that realise it ($P$), shown separately for the \textit{unrelated} (U) and \textit{coherent} (C) conditions. Reported \textit{t}-values come from Welch’s \textit{t}-tests comparing PO-$P$ and PO-$\overline{P}$ completions within each context. An asterisk indicates statistical significance after Benjamini--Hochberg FDR correction. All comparisons are statistically significant. This Table shows the PO-specific analysis of~\Cref{tab:rq3}.}
\label{tab:po_priming_similarity_tests}
\end{table*}

\begin{table*}[t]
\centering
\scriptsize
\setlength{\tabcolsep}{3.2pt}
\resizebox{\textwidth}{!}{
\begin{tabular}{ll|rrr|rrr|rrr|rrr|rrr|rrr|rrr}
\toprule
\textbf{Metric} & \textbf{Prim.}
& \multicolumn{3}{c|}{\textbf{Phi-2}}
& \multicolumn{3}{c|}{\textbf{Gemma-2-2B}}
& \multicolumn{3}{c|}{\textbf{Gemma-2-9B}}
& \multicolumn{3}{c|}{\textbf{EuroLLM-1.7B}}
& \multicolumn{3}{c|}{\textbf{EuroLLM-9B}}
& \multicolumn{3}{c|}{\textbf{Qwen-3-1.7B}}
& \multicolumn{3}{c}{\textbf{Qwen-3-8B}} \\
\cmidrule(lr){3-5}
\cmidrule(lr){6-8}
\cmidrule(lr){9-11}
\cmidrule(lr){12-14}
\cmidrule(lr){15-17}
\cmidrule(lr){18-20}
\cmidrule(lr){21-23}
& & \textbf{$\overline{P}$} & \textbf{$P$} & \textit{t}
  & \textbf{$\overline{P}$} & \textbf{$P$} & \textit{t}
  & \textbf{$\overline{P}$} & \textbf{$P$} & \textit{t}
  & \textbf{$\overline{P}$} & \textbf{$P$} & \textit{t}
  & \textbf{$\overline{P}$} & \textbf{$P$} & \textit{t}
  & \textbf{$\overline{P}$} & \textbf{$P$} & \textit{t}
  & \textbf{$\overline{P}$} & \textbf{$P$} & \textit{t} \\
\midrule

MiniLM & U
& \lightgraytext{0.25} & \lightgraytext{0.30} & \lightgraytext{2.29}
& 0.27 & 0.35 & 3.52
& 0.26 & 0.35 & 4.24
& 0.22 & 0.30 & 3.60
& 0.22 & 0.30 & 3.83
& \lightgraytext{0.27} & \lightgraytext{0.34} & \lightgraytext{3.18}
& \lightgraytext{0.26} & \lightgraytext{0.30} & \lightgraytext{2.01} \\
& C
& 0.34 & 0.45 & 6.84
& 0.37 & 0.47 & 6.61
& 0.39 & 0.55 & 10.38
& 0.30 & 0.40 & 6.17
& 0.35 & 0.49 & 9.06
& 0.36 & 0.46 & 5.99
& 0.39 & 0.53 & 10.04 \\

\midrule

MPNet & U
& \lightgraytext{0.26} & \lightgraytext{0.30} & \lightgraytext{1.97}
& \lightgraytext{0.27} & \lightgraytext{0.34} & \lightgraytext{2.97}
& 0.26 & 0.34 & 4.11
& 0.23 & 0.30 & 3.53
& \lightgraytext{0.22} & \lightgraytext{0.28} & \lightgraytext{3.36}
& 0.27 & 0.35 & 3.71
& \lightgraytext{0.26} & \lightgraytext{0.29} & \lightgraytext{1.50} \\
& C
& 0.33 & 0.42 & 6.53
& 0.36 & 0.44 & 5.84
& 0.37 & 0.50 & 9.12
& 0.30 & 0.38 & 5.13
& 0.34 & 0.45 & 8.14
& 0.35 & 0.43 & 5.11
& 0.37 & 0.49 & 8.74 \\

\midrule

BERTScore F1 & U
& \lightgraytext{0.19} & \lightgraytext{0.22} & \lightgraytext{2.21}
& \lightgraytext{0.16} & \lightgraytext{0.18} & \lightgraytext{1.34}
& \lightgraytext{0.17} & \lightgraytext{0.17} & \lightgraytext{0.20}
& \lightgraytext{0.14} & \lightgraytext{0.14} & \lightgraytext{0.19}
& \lightgraytext{0.16} & \lightgraytext{0.18} & \lightgraytext{1.99}
& \lightgraytext{0.14} & \lightgraytext{0.17} & \lightgraytext{2.40}
& \lightgraytext{0.14} & \lightgraytext{0.15} & \lightgraytext{0.90} \\
& C
& 0.25 & 0.29 & 4.68
& 0.19 & 0.23 & 3.92
& 0.20 & 0.27 & 7.48
& \lightgraytext{0.17} & \lightgraytext{0.20} & \lightgraytext{2.81}
& 0.19 & 0.24 & 5.39
& 0.17 & 0.22 & 4.17
& 0.19 & 0.26 & 7.87 \\

\midrule

Jaccard & U
& 0.06 & 0.14 & 5.07
& 0.07 & 0.14 & 5.35
& 0.07 & 0.16 & 6.98
& 0.05 & 0.15 & 5.98
& 0.06 & 0.14 & 7.11
& 0.07 & 0.16 & 5.90
& 0.07 & 0.14 & 5.39 \\
& C
& 0.12 & 0.28 & 12.12
& 0.12 & 0.25 & 10.19
& 0.14 & 0.33 & 13.29
& 0.09 & 0.22 & 10.68
& 0.13 & 0.31 & 14.08
& 0.13 & 0.26 & 9.82
& 0.15 & 0.34 & 15.12 \\

\midrule

BLEU & U
& 4.52 & 7.53 & 4.17
& 4.62 & 7.21 & 3.86
& 4.76 & 8.18 & 5.15
& 4.18 & 7.86 & 3.88
& 4.66 & 8.23 & 4.86
& 4.80 & 8.04 & 4.38
& 4.89 & 8.22 & 5.38 \\
& C
& 5.77 & 14.25 & 10.92
& 5.46 & 12.67 & 9.36
& 6.12 & 17.78 & 13.22
& 4.99 & 11.46 & 9.08
& 6.02 & 16.65 & 13.71
& 6.00 & 13.83 & 9.31
& 6.48 & 18.48 & 15.81 \\

\bottomrule
\end{tabular}
}
\caption{RQ3. DO-specific similarity analysis on the \textbf{full set of generations}. Mean similarity scores for completions that do not realise the primed DO structure ($\overline{P}$) and completions that realise it ($P$), shown separately for the \textit{unrelated} (U) and \textit{coherent} (C) conditions. Reported \textit{t}-values come from Welch’s \textit{t}-tests comparing DO-$p$ and DO-$\overline{P}$ completions within each context. The results in black indicate statistical significance after Benjamini--Hochberg FDR correction. \lightgraytext{Gray text} indicates insignificant differences. This table shows the DO-specific analysis of~\Cref{tab:rq3}.}
\label{tab:do_priming_similarity_tests}
\end{table*}

\section{Additional Human-Annotated Analysis}
\label{app:human-annotated-additional}

In this appendix, we report complementary analyses on the balanced hand-annotated sample introduced in~\Cref{sec:results}, with the aim of verifying whether the same qualitative patterns are visible when model completions are labelled manually rather than by the automatic classifier.

The hand-annotated set contains 2,700 completions, balanced across three models, three prime structures, and three contextual conditions. 
For each model, we include 100 examples for each combination of target structure (\textsc{po}, \textsc{do}, \textsc{non-dative}) and context condition (\textit{coherence}, \textit{unrelated}, and \textit{no context}). 
All analyses in this appendix use the manually assigned gold labels. 
Thus, a completion is treated as structurally primed when the gold label matches the prime structure.

We organise the appendix according to the three research questions addressed in the main analysis. 
First, we test whether unrelated dative primes increase the production of matching dative structures relative to baseline conditions. 
Second, we examine whether semantic coherence strengthens this priming effect. 
Third, we test whether structurally primed completions are also more closely aligned with their primes in semantic and lexical space.

\begin{table*}[t]
    \centering
    \small
    \setlength{\tabcolsep}{4pt}
    \renewcommand{\arraystretch}{1.15}
    \scalebox{0.9}{
    \begin{tabular}{p{1.3cm}p{6.0cm}p{6.8cm}p{1.4cm}}
        \toprule
        \textbf{Model} & \textbf{Prompt} & \textbf{LM completion} & \textbf{Label} \\
        \midrule

        EuroLLM 
        & A winner threw a juice to a husband. A husband kept the
        & A husband kept the juice and drank it.
        & non-dative \\

        EuroLLM 
        & A lawyer saved a tea for a buddy. A buddy gave the
        & A buddy gave the scruffy-looking solo lawyer a drink since they tasted funny.
        & DO \\

        EuroLLM 
        & A husband brought a meal to a writer. A writer designed the
        & A writer designed the meal by placing between each serving of meat and each vegetable a single small, leafy, green vegetable.
        & non-dative \\

        EuroLLM 
        & An attorney made a camera for an opponent. An opponent gave the
        & An opponent gave the attorney 10 yellow cameras.
        & DO \\

        \midrule

        Gemma 
        & A man promised a husband a coffee. A man purchased the
        & A man purchased the goods, but instead put coffee on sugar.
        & non-dative \\

        Gemma 
        & A husband purchased the
        & A husband purchased the house his wife grew up in.
        & non-dative \\

        Gemma 
        & A man purchased the
        & A man purchased the first 22 bottles of soda at 1.69 each and 34 bottles of soda at 1.
        & non-dative \\

        Gemma 
        & It was dangerous, but exciting. Angela sold the
        & Angela sold the group a lie that if they could travel to a place called the ``Lost Land'' by storm and get the key to a forbidden door that was sealed shut, then they would be transported to a futuristic...
        & DO \\

        \midrule

        Phi 
        & The prince saved the computer for the secretary. A professor gave a
        & A professor gave a paper to a student.
        & PO \\

        Phi 
        & The aunt bought the beer for the father. A colonel sent a
        & A colonel sent a formal casket to the father.
        & PO \\

        Phi 
        & A mayor bought a juice for an actor. The husband took the
        & The husband took the juice to the cinema in two parts.
        & non-dative \\

        Phi 
        & A student sent a pie to an army. An army drew the
        & An army drew the line between them.
        & non-dative \\

        \bottomrule
    \end{tabular}
    }
    \caption{Illustrative LM completions from the manual annotation process. The examples include cases where generated continuations superficially resemble dative constructions, but require manual inspection to determine whether they realise a genuine PO, DO, or non-dative structure.}
    \label{tab:ambiguous_lm_completions}
    \vspace{-5pt}
\end{table*}

\subsection{RQ1: Structural Priming Against Baseline Conditions}
\label{app:rq1_gold}

\begin{table*}[t]
\centering
\small
\begin{tabular}{llrrrrrrrrr}
\toprule
\textbf{Model} & \textbf{Target} & \textbf{No Context} & \textbf{Non-dative} & \textbf{Unrelated} 
& $\Delta^{N}$ & $z^{N}$ & $p^{N}_{\mathrm{BH}}$ 
& $\Delta^{O}$ & $z^{O}$ & $p^{O}_{\mathrm{BH}}$ \\
\midrule
EuroLLM-1.7B & PO & 6.00 & 14.00 & 15.00 & 9.00 & 2.08 & .114 & 1.00 & 0.20 & .841 \\
              & DO & 5.00 & 9.00  & 7.00  & 2.00 & 0.60 & .662 & -2.00 & -0.52 & .841 \\
\midrule
Gemma-2-2B   & PO & 11.00 & 8.00 & 19.00 & 8.00 & 1.58 & .226 & 11.00 & 2.28 & .137 \\
              & DO & 2.00  & 5.00 & 4.00  & 2.00 & 0.83 & .611 & -1.00 & -0.34 & .841 \\
\midrule
Phi-2-2.7B     & PO & 6.00 & 12.00 & 22.00 & 16.00 & 3.26 & .007 & 10.00 & 1.88 & .179 \\
              & DO & 8.00 & 5.00  & 9.00  & 1.00  & 0.25 & .800 & 4.00  & 1.11 & .535 \\
\bottomrule
\end{tabular}
\caption{
RQ1 results on the \textbf{hand-annotated} set. For each target structure, \textit{Unrelated} reports the percentage of structure-matching completions after unrelated dative primes. \textit{No Context} is the no-context baseline, and \textit{Non-Dative} is the non-dative-prime baseline. $\Delta^{N}$ and $z^{N}$ compare unrelated dative primes against the no-context baseline; $\Delta^{O}$ and $z^{O}$ compare them against the \textsc{non-dative}-prime baseline. $p$-values are Benjamini--Hochberg corrected.
}
\label{tab:rq1_annotated}
\end{table*}

We first ask whether exposure to an \textit{unrelated} dative prime increases the likelihood of producing the corresponding dative structure. 
As in~\Cref{subsec:rq1_baselines}, we compare the unrelated dative-prime condition against two baselines: the \textit{no context} condition and the \textsc{non-dative}-prime condition. 
For each model and target structure, we compute the proportion of completions whose gold label matches the target structure.

~\Cref{tab:rq1_annotated} shows that the hand-annotated sample provides only partial evidence for structural priming in unrelated contexts. 
For \textsc{PO}, all three models produce more matching completions after \textit{unrelated} \textsc{PO} primes than in the \textit{no-context} baseline. 
However, after correction, this contrast is statistically reliable only for Phi-2-2.7B. 
The same pattern is weaker when the \textit{unrelated} condition is compared against the \textsc{non-dative}-prime baseline: the differences are positive for all models, but do not reach significance after correction.

For \textsc{DO}, the \textit{unrelated}-prime effect is much smaller. 
All three models show only slight increases over the \textit{no-context} baseline, and none of the comparisons is statistically reliable. 
This suggests that, in the hand-annotated sample, the \textit{unrelated} structural priming is more visible for \textsc{PO} than for \textsc{DO}. 
This asymmetry is consistent with the broader tendency observed in the main experiments: \textsc{PO} completions are more frequent and easier to elicit, whereas \textsc{DO} completions are rarer and appear to depend more strongly on supportive contextual conditions.

Overall, the manually annotated results support the direction of the large-scale findings, but with weaker statistical evidence. 
This is expected given the smaller sample size of the gold-labelled set. 
The hand-annotated analysis therefore confirms that, since \textit{unrelated} primes can increase structure-matching production, (especially for \textsc{PO}), structural priming occurs. 

\subsection{RQ2: Coherence Effects on Structural Priming}
\label{app:rq2_annotated}

\begin{table*}[t]
\centering
\small
\begin{tabular}{lrrrrr|rrrrr}
\toprule
& \multicolumn{5}{c}{\textbf{PO priming}} 
& \multicolumn{5}{c}{\textbf{DO priming}} \\
\cmidrule(lr){2-6}\cmidrule(lr){7-11}
\textbf{Model} 
& \textbf{Coher.} & \textbf{Unrel.} & $\Delta^{C-U}$ & $z$ & $p_{\mathrm{BH}}$
& \textbf{Coher.} & \textbf{Unrel.} & $\Delta^{C-U}$ & $z$ & $p_{\mathrm{BH}}$ \\
\midrule
EuroLLM-1.7B & 20.00 & 15.00 & 5.00  & 0.93  & .423 & 21.00 & 7.00 & 14.00 & 2.85 & .013 \\
Gemma-2-2B   & 25.00 & 19.00 & 6.00  & 1.02  & .423 & 18.00 & 4.00 & 14.00 & 3.16 & .009 \\
Phi-2-2.7B     & 21.00 & 22.00 & -1.00 & -0.17 & .863 & 19.00 & 9.00 & 10.00 & 2.04 & .083 \\
\bottomrule
\end{tabular}
\caption{
RQ2 results on the \textbf{hand-annotated} set, comparing coherent and unrelated prime conditions. Values are percentages except for $z$ and corrected $p$-values. $\Delta^{C-U}$ denotes the difference between coherent and unrelated conditions. Reliability is assessed with two-sample proportion $z$-tests with Benjamini--Hochberg FDR correction.
}
\label{tab:rq2_gold_annotated}
\end{table*}

We next examine whether semantic coherence between the prime and the target stem strengthens structural priming. 
As in~\Cref{subsec:rq2_coherence_effects}, we compare the \textit{coherence} and \textit{unrelated} conditions while holding the target structure fixed. 
A positive difference indicates that coherent prime--target pairs lead to more structure-matching completions than \textit{unrelated} prime--target pairs.

~\Cref{tab:rq2_gold_annotated} shows that the \textit{coherence} effect differs across the two dative alternations. 
For \textsc{PO}, the effect is small and inconsistent. 
EuroLLM-1.7B and Gemma-2-2B show numerically higher \textsc{PO} production in the \textit{coherence} condition, whereas Phi-2-2.7B shows virtually no increase. 
None of the \textsc{PO} comparisons remains statistically reliable after correction.

The pattern is clearer for \textsc{DO}. 
Both EuroLLM-1.7B and Gemma-2-2B show a substantial and statistically reliable increase in \textsc{DO} completions under \textit{coherence}. 
Phi-2-2.7B follows the same direction, although the effect does not survive correction. 
This indicates that semantic coherence plays a more important role for \textsc{DO} priming than for \textsc{PO} priming in the hand-annotated sample.

Taken together, these results mirror the qualitative conclusion of the full-set analysis: coherence does not simply increase dative production uniformly, but appears to be especially important for the less frequent and more constrained \textsc{DO} alternation. 
The manually annotated data therefore provide a more conservative but compatible picture: \textsc{PO} priming can emerge even without strong semantic support, while \textsc{DO} priming is more dependent on coherent contextual cues.

\subsection{RQ3: Lexico-semantic Alignment under Structural Priming}
\label{app:rq3_gold_annotated}

\begin{table*}[t]
\centering
\scriptsize
\begin{tabular}{ll|rrr|rrr|rrr}
\toprule
\textbf{Metric} & \textbf{Cond.} 
& \multicolumn{3}{c|}{\textbf{EuroLLM-1.7B}} 
& \multicolumn{3}{c|}{\textbf{Gemma-2-2B}} 
& \multicolumn{3}{c}{\textbf{Phi-2-2.7B}} \\
& 
& \textbf{$\overline{P}$} & \textbf{$P$} & $t$ 
& \textbf{$\overline{P}$} & \textbf{$P$} & $t$ 
& \textbf{$\overline{P}$} & \textbf{$P$} & $t$ \\
\midrule
MiniLM & U 
& 0.28 & 0.31 & 0.64 
& 0.29 & 0.40 & 2.55$^{*}$ 
& 0.25 & 0.32 & 2.06 \\
& C 
& 0.36 & 0.38 & 0.62 
& 0.39 & 0.47 & 2.46$^{*}$ 
& 0.37 & 0.47 & 3.64$^{*}$ \\
\midrule

MPNet & U 
& 0.27 & 0.30 & 0.94 
& 0.28 & 0.36 & 1.75 
& 0.25 & 0.31 & 2.06 \\
& C 
& 0.34 & 0.37 & 0.79 
& 0.37 & 0.44 & 2.38$^{*}$ 
& 0.35 & 0.48 & 4.62$^{*}$ \\
\midrule

BERTScore F1 & U 
& 0.17 & 0.20 & 1.43 
& 0.18 & 0.23 & 2.56$^{*}$ 
& 0.21 & 0.26 & 2.44$^{*}$ \\
& C 
& 0.18 & 0.20 & 0.80 
& 0.21 & 0.23 & 1.13 
& 0.26 & 0.32 & 3.23$^{*}$ \\
\midrule

Jaccard & U 
& 0.08 & 0.11 & 2.16 
& 0.08 & 0.11 & 1.77 
& 0.07 & 0.12 & 2.72$^{*}$ \\
& C 
& 0.11 & 0.19 & 3.25$^{*}$ 
& 0.12 & 0.21 & 3.87$^{*}$ 
& 0.11 & 0.22 & 5.50$^{*}$ \\
\midrule

BLEU & U 
& 4.52 & 6.03 & 2.20 
& 4.42 & 4.81 & 0.91 
& 4.25 & 6.71 & 2.76$^{*}$ \\
& C 
& 4.96 & 9.03 & 3.60$^{*}$ 
& 5.17 & 8.78 & 3.42$^{*}$ 
& 5.02 & 9.55 & 5.49$^{*}$ \\
\bottomrule
\end{tabular}
\caption{
RQ3 results on the \textbf{hand-annotated} set. Mean similarity scores are reported for non-primed ($\overline{P}$) and primed ($P$) completions in the \textit{unrelated} (U) and \textit{coherent} (C) conditions. Reported $t$ values come from Welch's $t$-tests comparing primed and non-primed completions within each condition and model. Asterisks indicate significance after Benjamini--Hochberg FDR correction ($p_{\mathrm{BH}} < .05$).
}
\label{tab:rq3_gold_annotated}
\end{table*}

Finally, we examine whether structural priming in the hand-annotated set is accompanied by greater lexico-semantic alignment between the prime and the generated completion. 
As in~\Cref{subsec:rq3_semantic_evidence_full}, we compare structurally primed ($P$) and non-primed ($\overline{P}$) completions using five complementary metrics: MiniLM cosine similarity, MPNet cosine similarity, BERTScore F1, Jaccard lexical overlap, and BLEU. 
Similarity scores are computed only for the \textit{coherence} and \textit{unrelated} conditions, since the \textit{no context} condition does not provide a prime sentence against which the completion can be compared.

~\Cref{tab:rq3_gold_annotated} shows that the relation between structural priming and lexico-semantic alignment is present, but less uniform than in the large-scale automatic analysis. 
The clearest pattern is observed for Phi-2-2.7B. 
In the \textit{coherence} condition, primed completions obtain higher scores than non-primed completions across all metrics, indicating that structural repetition is accompanied by both semantic continuity and lexical reuse. 
In the \textit{unrelated} condition, the same tendency is visible but less consistent, with stronger effects for surface-oriented measures such as Jaccard overlap and BLEU.

Gemma-2-2B shows a similar but more metric-dependent pattern. 
Primed completions tend to be more similar to their primes than non-primed completions, particularly under coherence. 
However, the effect is less stable across metrics than for Phi-2-2.7B, suggesting that structural repetition in Gemma-2-2B is associated with lexico-semantic alignment, but not uniformly across all similarity measures.

EuroLLM-1.7B displays the weakest alignment pattern. 
The semantic encoders show only limited separation between primed and non-primed completions, whereas lexical metrics show clearer increases, especially in the coherence condition. 
This suggests that, for EuroLLM-1.7B, manually identified structural priming is more strongly associated with surface-level lexical reuse than with broader sentence-level semantic similarity.

Overall, the hand-annotated analysis supports the main conclusion of RQ3, but in a more nuanced form. 
Structurally primed completions are often more semantically and lexically aligned with their primes, especially in coherent contexts. 
However, the strength of this coupling varies by model and by metric. 
The gold-labelled sample therefore confirms that structural priming is not purely syntactic, while also showing that the association between structural repetition and lexico-semantic alignment is weaker and more variable when evaluated on a smaller manually annotated set.

% ------------------------------------------------------ OLD template
% \section{Example Appendix}
% \label{sec:appendix}

% This is a section in the appendix.

\end{document}